\documentclass[lettersize,journal]{IEEEtran}
\usepackage{amsmath,amsfonts}
\usepackage{algorithmic}
\usepackage{algorithm}
\usepackage{array}
\usepackage[caption=false,font=normalsize,labelfont=sf,textfont=sf]{subfig}
\usepackage{textcomp}
\usepackage{stfloats}
\usepackage{url}
\usepackage{verbatim}
\usepackage{graphicx}
\usepackage{cite}
\usepackage{color}
\usepackage{adjustbox}
\usepackage{hyperref}
\usepackage[capitalize,noabbrev]{cleveref}
\usepackage{booktabs} 
\usepackage{multirow}
\usepackage{enumitem}

\def\vlfull{{Vision-Language}}

\def\vlm{{VLM}}
\def\clcfull{{Compositional Language Concepts}}
\def\clc{{CLC}}

\def\oursm{3VL}

\def\oursmfull{{\textbf{Tree}-augmented \textbf{V}ision-\textbf{L}anguage}}

\def\oursi{DiRe}
\def\oursifull{{\textbf{Di}fferential
\textbf{Re}levance}}

\newcommand{\rev}[1]{{\textcolor{black}{#1}}}

\NewDocumentCommand{\raja}{ mO{} }{\noindent\textcolor{purple}{\textsf{\small[#1]}\textsuperscript{\textit{Raja}}}}

\NewDocumentCommand{\nir}{ mO{} }{\noindent\textcolor{red}{\textsf{\small[#1]}\textsuperscript{\textit{Nir}}}}

\NewDocumentCommand{\leonid}{ mO{} }{\noindent\textcolor{green}{\textsf{\small[#1]}\textsuperscript{\textit{Leonid}}}}

\definecolor{cyan}{rgb}{0.45,0.87,0.95}
\definecolor{orange}{rgb}{0.95,0.8,0.6}
\definecolor{britishracinggreen}{rgb}{0.0, 0.26, 0.15}
\definecolor{cadmiumgreen}{rgb}{0.0, 0.42, 0.24}

\definecolor{citecolor}{RGB}{34,139,34}
\definecolor{lightred}{RGB}{241,140,142}

\begin{document}

\title{3VL: Using Trees to Improve \\ Vision-Language Models' Interpretability}

\author{Nir Yellinek, Leonid Karlinsky and Raja Giryes
\thanks{Manuscript received November 28, 2023}}

\markboth{}%
{Yellinek \MakeLowercase{\textit{et al.}}:Generative Adversarial Encoder Learning} 


\maketitle

\begin{abstract}
\vlfull{} models (\vlm{}s) have proven to be effective at aligning image and text representations, producing superior zero-shot results when transferred to many downstream tasks. However, these representations suffer from some key shortcomings in understanding \clcfull{} (\clc{}), such as recognizing objects' attributes, states, and relations between different objects. Moreover, \vlm{}s typically have poor interpretability, making it challenging to debug and mitigate compositional-understanding failures. In this work, we introduce the architecture and training technique of \oursmfull{} (\textbf{\oursm{}}) model accompanied by our proposed 
\textbf{Anchor} inference method and \oursifull{} 
(\textbf{\oursi{}}) interpretability tool. By expanding the text of an arbitrary image-text pair into a hierarchical tree structure using language analysis tools, \oursm{} allows the induction of this structure into the visual representation learned by the model, enhancing its interpretability and compositional reasoning. Additionally, we show how Anchor, a simple technique for text unification, can be used to filter nuisance factors while increasing \clc{} understanding performance, e.g., on the fundamental VL-Checklist benchmark. We also show how \oursi{}, which performs a differential comparison between \vlm{} relevancy maps, enables us to generate compelling visualizations of the reasons for a model's success or failure. Our code is available at:  \href{https://github.com/niryellinek/3VL}{https://github.com/niryellinek/3VL}.  
\end{abstract}

\section{Introduction}
\label{chapter:int}

In recent years, \vlfull{} models (\vlm{}s) have emerged as a powerful tool for tasks such as image captioning, visual question answering, and image retrieval. These models have achieved remarkable success owing to their ability to align visual and textual features and extract meaningful information from them. However, these models have several limitations in understanding \clcfull{} (\clc{}) such as recognizing objects' attributes, states, and relations between different objects. While VLMs have shown remarkable performance in zero-shot transfer learning, their results are not interpretable and they typically struggle with compositional reasoning, where they need to understand the relationships between different concepts in a sentence to provide an accurate response.

Previous works on improving compositionality in \vlm{}s focused mostly on augmenting data points in text space \cite{doveh2023teaching,hao2022mixgen,cascantebonilla2023going,ray2019sunny,tang2020semantic}. While this approach has led to some performance improvements, it offers no extra interpretability. Several approaches have been proposed for interpreting neural networks ranging from attribution methods to visualization techniques \cite{simonyan2013deep,Selvaraju2017GradCAM,ribeiro2016should,sundararajan2017axiomatic,lundberg2017unified,Kolek2022Cartoon,kolek2023explaining}. However, most are designed to explain an already trained network, and interpretability is often limited to a local and task-specific context \cite{adebayo2018sanity,yang2019benchmarking,Kindermans2022Unreliability,shah2021input,slack2020fooling,rudin2019stop,koh2020concept,subramanya2019fooling}. Thus, it has been suggested that a better approach is explainability by design \cite{schwalbe2023comprehensive}, i.e., interpretability is built into the model architecture and training process. By doing so, a more general and systematic interpretation and analysis of the model is enabled. This makes it easier to identify failure modes and biases early in the development process. Furthermore, such an approach provides visualizations that can help users understand and analyze the model's behavior, making it more transparent and trustworthy. In this paper, we take a step towards facilitating such explainability for \vlm{}s.

We propose a novel approach to address \vlm{}s' limitations in compositional understanding and interpretability by leveraging natural language hierarchical structures. Our approach is based on the idea of expanding the text of an image-text pair into a hierarchical tree structure using language analysis tools and inducing this structure into the visual representation learned by the model. Specifically, we propose the \oursmfull{} (\oursm{}) model architecture and training technique, which allows for a rich exploration of the text space using several levels of incremental text augmentation from coarse to fine-grained.
Each level of the tree represents a progressive refinement of the text structure and captures an increasingly detailed aspect of the image-text relationship. We use this hierarchical structure to guide the learning of visual features in a way that improves the model's compositional reasoning ability.




To complement our tree-based structured method, we introduce two novel inference and interpretability techniques called Anchor and \oursifull{} (\oursi{}). These approaches extract image relevancy maps using HilaCAM \cite{HilaCAM} based on positive and negative texts. By comparing the relevancy maps of images with positive text to that of the same image with negative text, these strategies can identify further weaknesses of \vlm{}s and filter out the noise and irrelevant information from the input. This leads to an increase in \vlm{}s' performance on \clcfull{}{} tasks. Furthermore, our proposed techniques also provide insights into the underlying failure modes of the used models.

We evaluate our approach on popular \clcfull{} benchmarks and demonstrate that our method achieves state-of-the-art performance, outperforming existing \vlm{}-based methods on these tasks. Furthermore, we show that our approach is not only effective but also interpretable, enabling us to generate compelling visualizations of the reasons for model success or failure and effectively filter the image signal for increased performance.

By explainable in the paper we mean that our model emphasizes better the more important parts in the image, where the importance is correlated to human judgment. Thus, to assess that our model is more explainable according to this definition, we use token removal that checks the importance of patches remaining in the image and a user study that checks the correlation between what is important to humans and what is important to the model.


This paper is organized as follows. In \cref{chapter:int} we briefly describe the topics related to this research and the main goals of this work.
In \cref{chapter:background}, we provide the necessary background and review the existing literature on related topics. 
In \cref{sec:3vl}, we present our novel tree-based training technique and inference method. This approach improves the interpretability of \vlm{}s. In \cref{sec:removal}, we provide additional interpretability tools. In \cref{chapter:exp}, we conduct experiments to test the effectiveness of our proposed methods. Finally, in \cref{chapter:sum}, we summarize our contributions and propose some ideas for possible future work.

\begin{figure*}[t]
\centering
\begin{adjustbox}{width=\textwidth}
\includegraphics{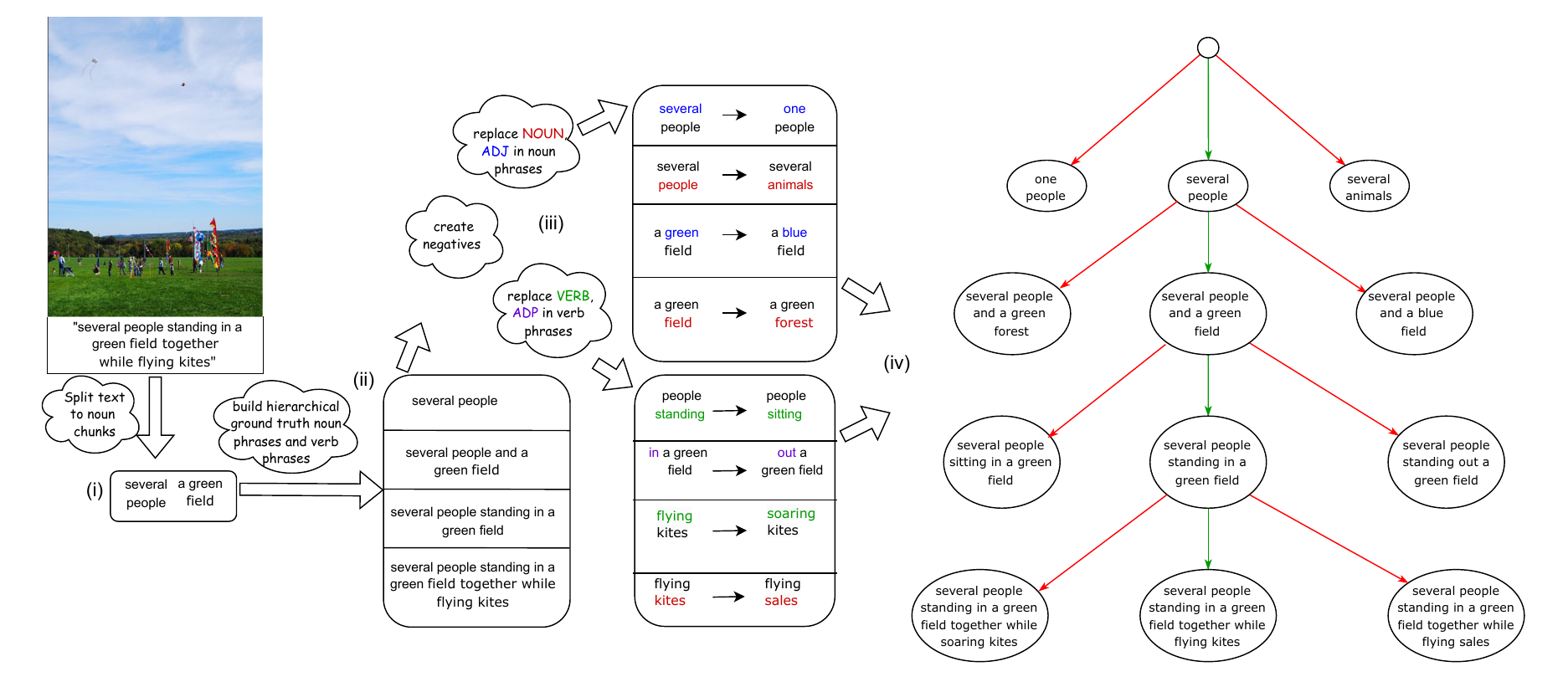}
\end{adjustbox}
\caption{The caption tree generation flow: (i) parse the sentence to get noun phrases and part of speech (ii) hierarchically reconstruct the caption (iii) generate negatives for each sub-caption (iv) compose the final tree.}
\label{fig:tree}
\vspace{-0.1in}
\end{figure*}

\section{Background}\label{chapter:background}

This section begins with a review of existing methods for the interpretability of deep neural networks and then discusses the usage of trees for deep learning and how our approach differs from previous works. Finally, it describes existing efforts for compositional understanding in \vlm{}s.

\subsection{Interpretability of deep neural networks}
Many approaches have been developed to explain the decisions of neural networks
\cite{Zhang2021SurveyInterpretability,schwalbe2023comprehensive}. Most works interpret decisions of an already trained network. Gradient Saliency \cite{simonyan2013deep}, GradCAM \cite{Selvaraju2017GradCAM}, and
Integrated Gradients \cite{sundararajan2017axiomatic} use convolutional neural network gradients to find significant pixels in the input that affect the network output. LIME \cite{ribeiro2016should} and SHAP \cite{lundberg2017unified} segment first the input into superpixels to find regions that explain the output. The work in \cite{Chiquier2024Evolving} use large language models (LLMs) to interpret \vlm{}s.
Wavelets \cite{Kolek2022Cartoon} and shearlets \cite{kolek2023explaining} have been used to further improve region selection. To provide similar interpretability tools for transformers, Chefer et al. \cite{Chefer_2021_CVPR} proposed to use the concept of layer-wise Relevance Propagation \cite{bach2015pixel}, with gradient integration for the self-attention layers of transformers. HilaCAM \cite{HilaCAM} proposed a more general method to visualize transformer architectures more easily. \rev{This method generates a relevancy map for each interaction between input modalities, whether through self-attention or bi-modal attention. Each relevancy map is derived from a forward pass through the attention layers and is averaged across the attention heads of each map using gradients.}

Another line of work focuses on `explainability by design' pointing to some flaws \cite{adebayo2018sanity,yang2019benchmarking,Kindermans2022Unreliability,shah2021input,slack2020fooling,rudin2019stop,koh2020concept,subramanya2019fooling}
in existing solutions for neural networks that are already trained. For example, \cite{adebayo2018sanity} shows that some saliency methods are independent of the model parameters and training data and thus cannot explain the failures of the models.
Instead, it is suggested that interpretable features are incorporated into the structure of the model.
One approach generates a set of linear transformations that are applied to the input \cite{bohle2021convolutional}. Another strategy encourages the network to make decisions that are locally linear \cite{alvarez2018towards}. This linearity makes the decision more interpretable. Another method \cite{Chattopadhyay2023Interpretable} generates a sequence of queries and makes a decision based on them. In \cite{chattopadhyay2023variational} a variational model is used to optimize the queries usage. An alternative approach is to learn interpretable semantic concepts from the data and then make a decision based on them \cite{yeh2020completeness,donnelly2022deformable,nauta2021neural,sarkar2022framework,lindner2023tracr}. A similar technique teaches the network to output also a `why' concept in addition to its decision \cite{Chengzhi2023Doubly}. StarNet~\cite{karlinsky2021starnet} incorporates a star model into network learning to get a relationship between locations in the query image and the decision of the network. \rev{The back projection generates decision evidence heatmaps, effectively explaining the model’s decision.
Our work differs from the above by incorporating both compelling visualizations of model decisions using HilaCAM and explainability by design using a fine-grained captions tree.}

\begin{figure*}[t]
\centering
\begin{adjustbox}{width=0.95\textwidth}
\includegraphics{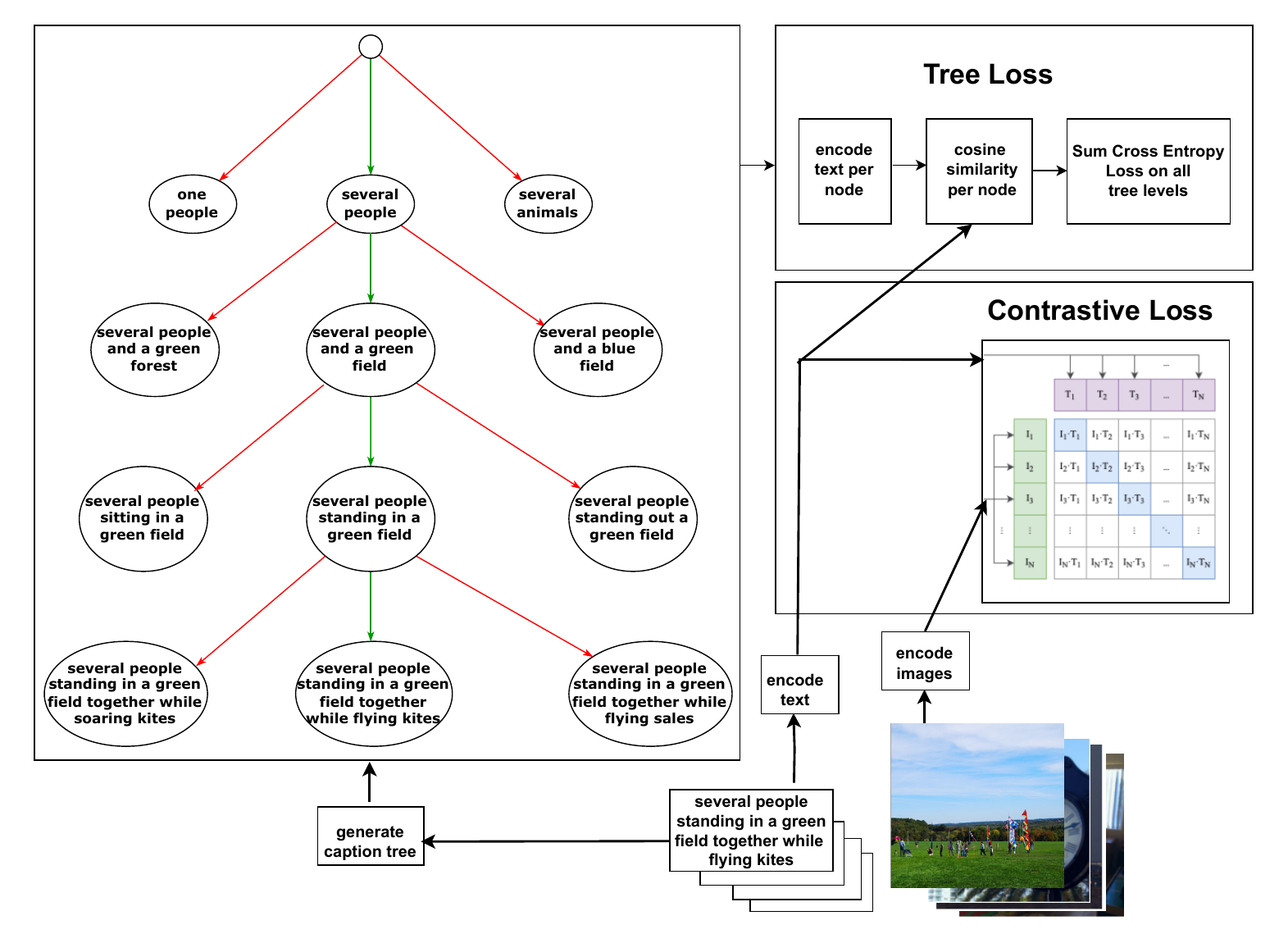}
\end{adjustbox}
\vspace{-0.12in}
\caption{The tree loss and contrastive loss that are used for training \oursm{}. For the tree loss we first generate a caption tree and then sum the cross entropy loss in all tree levels. For the contrastive loss, we calculate the average cross-entropy loss over all image-text pairs in the batch.}
\label{fig:training}
\vspace{-0.1in}
\end{figure*}

\subsection{Tree usage for deep learning}
Trees have been used with neural networks in different constellations \cite{li2022survey}. They have been incorporated into LSTMs \cite{tai2015improved} for improving semantic representations. Tree LSTMs have been combined with dependency parsing trees (DPTs) to improve visual grounding \cite{liu2019learning}. 
In \cite{Kontschieder_2015_ICCV,Tanno2019Adaptive} decision trees were combined with the representation learning functionality of neural networks. In \cite{wan2022unsupervised} trees were used with \vlm{}s for improving their grammar induction in an unsupervised manner.

\rev{To improve the network's explainability, a random forest has been trained on the network data. Its decisions were used to regularize the neural network such that it generates axis-aligned decisions \cite{wu2021optimizing}. Random forests have also been used in conjunction with pre-trained neural networks to improve the latter's robustness to adversarial attacks} \cite{ding2019defending,cohen2021simple}. 
Another strategy that uses trees to improve interpretability is NBDT \cite{wan2021nbdt} which replaces the neural network final layer with a decision tree, which helps to understand better the neural network mistakes. In \cite{alper2024hierarcaps} the hierarchical representations of \vlm{}s are analyzed and a method has been proposed to improve them.

\rev{Our work uses a tree both for achieving explainability by design and generating several levels of incremental text augmentations.} The tree decomposes the sentences in a structured manner, which intuitively should help the network better understand the input structures and therefore make them more interpretable and easier to decompose. Moreover, our structuring improves the representations and grounding also with our negative augmentation method, which is much richer than previous works of negative augmentation. \rev{Note also that our tree structure helps inspecting VLM’s failures, e.g., as we show in Section~\ref{sec:failure_cases}.}

\subsection{Compositionality in \vlfull{} Models}
One of the key challenges in developing effective \vlm{}s is achieving a compositional understanding of the underlying visual and linguistic elements.

Some \vlm{}s such as CLIP~\cite{clip} and ALIGN~\cite{align} have shown great progress in zero-shot downstream tasks by pre-training on a large-scale dataset of text-image pairs collected from the web using contrastive image-text alignment. Other approaches like LXMERT~\cite{tan2019lxmert}, UNITER~\cite{chen2020uniter}, and OSCAR~\cite{li2020oscar} utilize off-the-shelf object detectors to extract region features.

More recent works have explored techniques for fine-grained contrastive learning and additional positives from nearest neighbors to improve image-text retrieval on benchmarks such as ImageNet~\cite{russakovsky2015imagenet} and MS-COCO~\cite{lin2014microsoft}. These models include FILIP~\cite{yao2021filip}, CyClip~\cite{cyclip}, DeCLIP~\cite{declip}, and PyramidCLIP~\cite{gao2022pyramidclip}.

Some recent works propose self-supervised learning objectives such as image-text matching and masked/autoregressive language modeling to improve the model's ability to understand vision and language concepts. These works include VILT~\cite{kim2021vilt}, ALIGN~\cite{align}, Vision-TALK~\cite{yang2022vision}, ViCHA~\cite{shukor2022efficient} and BLIP~\cite{blip}. BLIP, for example, generates synthetic captions from the language modeling head and filters noisy captions based on the image-text matching score.

Despite these advances, recent studies like VL-CheckList~\cite{vlc}, the Winoground Challenge~\cite{winoground}, VSR \cite{Liu2022VisualSR}, VALSE \cite{Parcalabescu_2022} and COLA \cite{ray2023cola} have shown that \vlm{}s still struggle with understanding fine-grained language details and CLC. In VL-Checklist~\cite{vlc} we have a set of images and a pair of positive and negative captions for each image. The model is tested for image-text retrieval. This benchmark aims to evaluate \vlm{}s' compositional understanding of objects, attributes, and relations. Each negative caption replaces one word for Objects, Relations, or Attributes in the positive sentence. Objects are tested for size and location invariance, the Attributes category checks for recognition of five different attributes: color, material, size, action, and state, and the Relations category checks for recognition of two relations between objects: action relation and spatial relation. 
Another work has shown that there is a fundamental challenge for transformers with compositionality \cite{dziri2023faith}. Therefore, there is still much work to be done in developing \vlm{}s with stronger compositionality and robustness to complex, real-world scenarios.

Achieving a full understanding of the semantics of rich visual scenes involves the ability to detect individual entities, reason about their interactions and attributes, and ultimately understand the visual concepts within the scene. Structured representations have played a vital role in this process, and have been applied to a wide range of computer vision applications, such as vision and language~\cite{chen2020uniter,Li2019VisualBERTAS,li2020oscar,tan2019lxmert}, scene graphs~\cite{sg_generation_msg_pass,herzig2018mapping,referential_relationships,Jerbi2020LearningOD,raboh2020dsg}, relational reasoning~\cite{baradel2018object,battaglia2018relational}, human-object interactions~\cite{Gao2020DRGDR,Kato2018CompositionalLF,Xu2019LearningTD}, action recognition~\cite{avraham2022svit,arnab2021unified,materzynska2019something,herzig2022orvit,herzig2019stag,ji2019action,Wang_videogcnECCV2018}, and even image and video generation from graphs~\cite{2020ActionGraphs,herzig2019canonical,johnson2018image}. Some works~\cite{andreas2017neuralmodulenetworks,andreas2016learningcomposeneuralnetworks,hu2017learningreasonendtoendmodule} take advantage of the compositional nature of natural language for solving VQA.
However, most of these works rely on detailed, manually curated supervision that often involves the annotation of location information and structural details. This results in limited-size or synthetic data sources for training, which can limit the effectiveness of the model. In \cite{doveh2023teaching}, the authors have proposed a method for teaching CLC to large \vlm{}s using only the available large-scale noisy \vlfull{} data sources collected from the web, without the use of expensive manual curation. This approach aims to improve the scalability and effectiveness of \vlm{}s' understanding of \clcfull{}.

Perhaps most similar to our augmentation method is the work of \cite{doveh2023teaching}, which generates one negative caption per example by replacing one random word out of all possible candidates. In \cite{doveh2023teaching} the authors prepare in advance a closed set of words for each VL-Checklist category. The possible candidates for replacement are all words that match one of the categories' set of words. The negative word is chosen randomly from the corresponding set of words excluding the positive word itself.
Our proposed approach differs from the above by generating several augmentations per caption in the form of hierarchical trees for \vlm{} training and the use of WordNet and a LLM for the generation of each negative sample. Note that unlike previous works on compositional reasoning in \vlm{}s, our approach aims at improving both compositional reasoning and interpretability.

\section{The \oursmfull{} (\oursm{}) model}
\label{sec:3vl}

This section describes {\oursm{}}, our novel tree-based model architecture and training technique, and our novel {\textit{Anchor}} inference method and \oursifull{} ({\textit{\oursi{}}}) interpretability tool. First, we present our tree augmentation technique. Then, we discuss the details of our tree-based training method. In \cref{sec:removal}, we present our {\textit{Token Removal}} inference and interpretability tools.  

\subsection{Caption tree generation}

We present below our tree augmentation method. Figure \ref{fig:tree} illustrates this process.

\renewcommand{\labelenumii}{\arabic{enumi}.\arabic{enumii}}
\renewcommand{\labelenumiii}{\arabic{enumi}.\arabic{enumii}.\arabic{enumiii}}
\renewcommand{\labelenumiv}{\arabic{enumi}.\arabic{enumii}.\arabic{enumiii}.\arabic{enumiv}}

\begin{enumerate}[leftmargin=*]
    \item For each image caption pair we first parse the caption using \cite{spacy2} to get all noun phrases and part of speech tags.
    \item Then, we reconstruct the full caption hierarchically from coarse to fine-grained to get a positive sub-caption for each level in the tree in the following way: 

    (using the caption {\textit{``several people standing in a green field together while flying kites"}} as an example)

    \begin{enumerate}
    \item The first level of the tree will contain the first noun phrase as its positive text (i.e. {\textit{``several people"}}).
    \item The second level of the tree will contain the text of the first and second noun phrases concatenated with some connecting word like 'and' (i.e. {\textit{``several people and a green field"}}).
    \item The third level of the tree will contain the text of the original caption from the start until the end of the second noun phrase (i.e. {\textit{``several people standing in a green field"}}).
    \begin{enumerate}
        \item If more noun phrases exist in the original caption then in a similar way the next levels of the tree will contain the text of previous noun phrases concatenated to the current noun phrase with a word like 'and', and the original caption from the start until the end of the current noun phrase.
    \end{enumerate}
    \item Finally, the last level of the tree will contain the text of the full original caption (i.e. {\textit{``several people standing in a green field together while flying kites"}}).
    
    \end{enumerate}

\item Next, in each tree level we generate one negative caption for each Noun, Adjective, Adposition, and Verb of the positive text (each negative replaces just one word in the original caption).
Note that we do not replace again words that appeared in previous tree levels. So information from a previous level flows without change.
Each negative word is generated as follows:
\begin{enumerate}
    \item Find an opposite (Antonym) of the positive word using FLAN-T5 LLM \cite{chung2022scaling} with prompt (e.g. ``find an opposite for the word: $<>$").
    \item If an opposite is not found, we generate a co-hyponym\footnote{Co-hyponyms are words that share the same hypernym in the wordnet tree (e.g. "apple" and "banana" are co-hyponyms as they share the hypernym ``fruit"; the words ``car" and ``motorcycle", ``blue" and ``yellow" are co-hyponyms as well.} of the positive word using NLTK's \cite{bird2009natural} WordNet  \cite{miller1995wordnet} module.
    
    \item If a co-hyponym is not found, we use T5 LLM \cite{raffel2020exploring} to generate a word to fill in a masked positive word (the token '$<$extra\_id\_0$>$' replaces the positive word in the prompt).
\end{enumerate}
\end{enumerate}

For the above example caption we generate the following negative captions:
\begin{itemize}[leftmargin=*]
    \item At the first level we generate, for the positive text {\textit{``several people"}}, the negative texts  {\textit{``one people"}} and {\textit{``several animals"}}.
    \item At the second level we generate, for the positive text {\textit{``several people and a green field"}}, the negative texts {\textit{"several people and a blue field"}} and  {\textit{``several people and a green forest"}}.

    \item At the third level we generate, for the positive text {\textit{``several people standing in a green field"}}, the negative texts {\textit{``several people sitting in a green field"}} and  {\textit{``several people standing out a green field"}}.

    \item At the fourth level we generate, for the positive text {\textit{``several people standing in a green field together while flying kites"}}, the negative texts {\textit{``several people standing in a green field together while soaring kites"}} and {\textit{``several people standing in a green field together while flying sales"}}.
    
\end{itemize}

    Note that our automated negatives generation method can generate grammatical errors sometimes.



\subsection{Tree-based training}
\textbf{CLIP contrastive loss.} Given a batch of \(N\) image-caption pairs we extract the image and text representations, \(I_{r}\) and \(T_{r}\), using CLIP's image and text encoders. Then,  we compute the pairwise cosine similarity scores \(S_{j,k}\) for each image \(j\) and caption \(k\) in the batch. From the similarities matrix, we calculate the cross entropy loss with softmax over the rows 
(\( \mathcal{L}_{img} \)) and the cross entropy loss with softmax over the columns (\( \mathcal{L}_{txt} \)). The final CLIP contrastive loss is the average of these two losses: 
\begin{align}
\label{eq:contrast_loss}
    \mathcal{L}_{contrast} = \frac{\mathcal{L}_{img}+\mathcal{L}_{txt}}{\mathit{2}}.
\end{align}

\textbf{Tree-based loss.} 
For each image-caption pair, we first create a caption tree. Then, for each level of the tree, we calculate the cosine similarity scores between the image and all captions at that level and calculate the Cross Entropy Loss. The final tree loss, \( \mathcal{L}_{tree} \), is the sum of losses over all tree levels (Fig. \ref{fig:training}).

To preserve as much of CLIP's zero-shot capabilities we also include \( \mathcal{L}_{contrast} \), on the original MS-COCO dataset \cite{lin2014microsoft} (without extra negatives). Our final loss function is: 
\begin{align}
\label{eq:total_loss}
\mathcal{L}_{total}  =  \alpha\cdot \mathcal{L}_{tree} + (\mathit{1} -\alpha)\cdot \mathcal{L}_{contrast},
\end{align}
where $0 < \alpha < 1$ is a hyperparameter. We found $\alpha=0.5$ to work the best.

To further diminish zero-shot forgetting, we also employ LoRA \cite{hu2022lora}, following the work of \cite{doveh2023teaching} and train only the LoRA adapters while the base CLIP model parameters remain frozen.

\textbf{Training Details.}
We finetune OpenAI CLIP \cite{clip} ViT-B/32 \cite{dosovitskiyimage} with rank=1 LoRA adapters on the training set of MS-COCO \cite{lin2014microsoft} for 12 epochs.
\rev{We use an AdamW optimizer with a learning rate of $3e-6$ and weight decay $0.1$ and train with a batch size of 64 on a single GeForce RTX 2080 Ti NVIDIA GPU. We performed a hyperparameter sweep and chose the final parameters and the number of epochs based on the MS-COCO validation set.}


\section{Relevancy Maps based Token Removal and Interpretability} \label{sec:removal}

\subsection{\textit{Token Removal}}

{\textit{Token Removal}} removes the least significant image tokens according to a given relevancy map. \rev{We use a $7\times 7$ relevancy map which is generated by HilaCAM and holds a significance score per patch in the image. \rev{Thus, the image is divided also into $7\times 7$ equal-size patches,} where each patch is an image token.}
The image with the removed tokens is used as input to the image encoder. We call this approach HilaCAM with {\textit{Token Removal}.

Using the relevancy maps generated for each image-text pair and \textit{Token Removal} we point the model towards the more important parts of the image. 
By taking into account the relevancy maps of an image paired with a positive text and the same image paired with a negative text simultaneously, we are able to understand models' decisions better. Combined with {\textit{\oursm{}}} negatives tree generation we gain valuable insights into the underlying failure modes. See Section~\ref{sec:failure_cases} for more details.


\begin{figure}[t]
\centering
\begin{adjustbox}{max width=\columnwidth}
\includegraphics{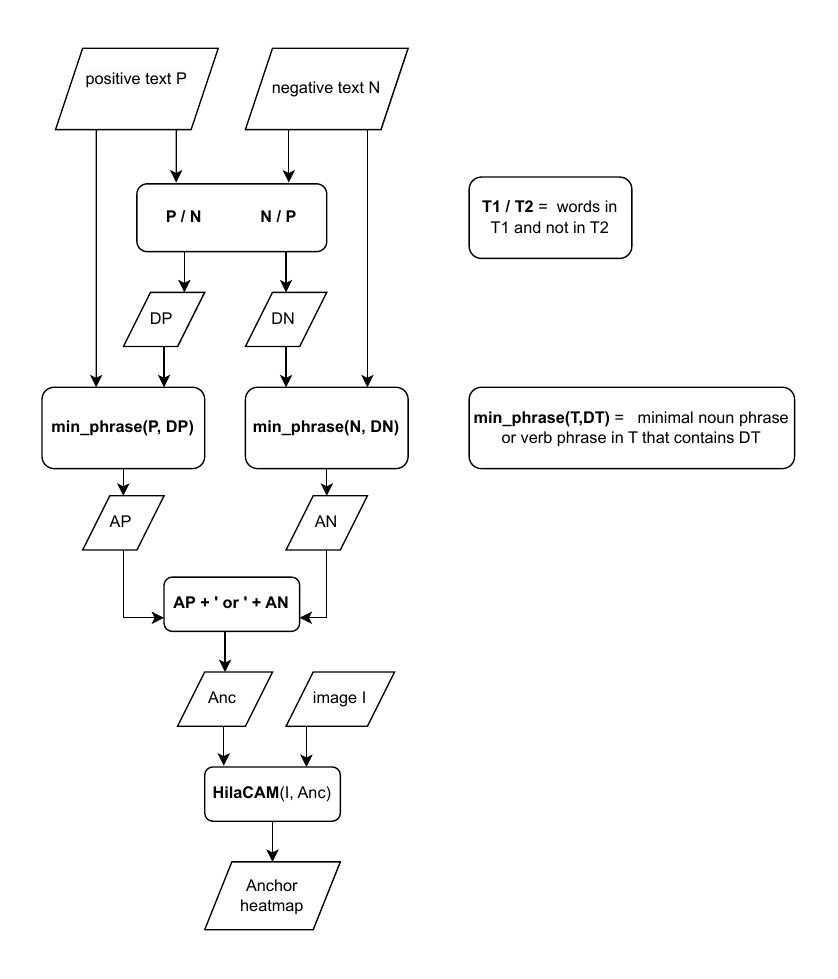}
\end{adjustbox}
\caption{Generating one relevancy heatmap using an ``Anchor'' text from two text possibilites. Note that unlike Figure~\ref{diag:two_texts_HilaCAM}, instead of having two heatmaps, here we have only one heatmap that is generated from the "Anchor" text that we create from the positive and negative texts.}
\label{diag:Anchor_diagram}
\end{figure}

\begin{figure}[t]
\centering
\begin{adjustbox}{max width=\columnwidth}
\includegraphics{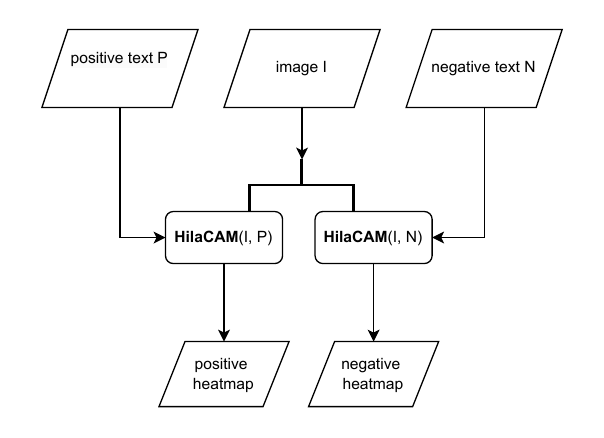}
\end{adjustbox}
\caption{\rev{When we have two possible texts for one image, we may apply HilaCAM two times to get two different relevancy heatmaps.}}
\label{diag:two_texts_HilaCAM}
\end{figure}

\subsection{\textit{HilaCAM Anchor}}
HilaCAM \cite{HilaCAM} generates a relevancy heatmap for a given image-text pair. In cases where two text \rev{possibilities} exist (e.g., one positive and one negative) for a single image, it is possible to generate one relevancy heatmap for each text paired with the given image. 
\rev{Formally, given an Image I, a positive text P, and a negative text N we apply HilaCAM two times to get two different relevancy heatmaps} as illustrated in Figure~\ref{diag:two_texts_HilaCAM}. This provides us with the following two heatmaps: 
\begin{align}
\label{eq:hila_cam}
positive\_heatmap = HilaCAM(I, P)\\
negative\_heatmap = HilaCAM(I, N)
\end{align}
By abuse of notation, we refer from now on to this natural way of calculating a different relevancy map per text as HilaCAM as well.
Each such relevancy map can be used for a process we call ‘Token Removal’ that is described hereafter.


Instead of using these two heatmaps, we propose another way to use HilaCAM in cases with a single image and two texts, e.g., one positive and one negative. In our approach, denoted as “Anchor”, we form a new single text from the two input texts and use this new text to generate a single relevancy map. 
In cases where the negative text was formed by replacing a few words in the positive text, it is possible to form a new single text by focusing on parts of texts that differ from one another. We call this new text as the “Anchor” text.

\rev{Formally, given an Image I, a positive text P, and a negative text N (that was generated by replacing a few words in P). We apply Anchor on P and N to get the new “Anchor” text Anc. Then we apply HilaCAM on Anc to get a single Anchor relevancy heatmap.} Figure~\ref{diag:Anchor_diagram} illustrates the use of HilaCAM with Anchor.
The following series of formulas describe how we get the anchor heatmap:\\
\textit{DP = P $\backslash$ N (words in P but not in N)}\\ 
\textit{DN = N $\backslash$  P (words in N but not in P)}\\
\textit{AP = minimal noun or verb phrase in P that contains DP}\\ 
\textit{AN = minimal noun or verb phrase in N that contains DN}\\
\textit{Anc = Anchor (P, N) = AP + ‘ or ‘ + AN}\\ 
\textit{anchor\_heatmap = HilaCAM(I, Anc)}\\

For example, given the two texts, {\textit{``people playing with airborne frisbee"}} and {\textit{``people playing with sitting frisbee"}} we can generate the new text {\textit{``airborne frisbee or sitting frisbee"}}. We refer to this new text as the {\textit{``Anchor"}} text. 

Having this {\textit{Anchor}} text, we can generate a relevancy map from it and use it for {\textit{Token Removal}}.  
{\textit{Anchor}} with {\textit{Token Removal}} is meant to focus both positive and negative texts on the same parts of the image. These parts of the image should contain the most important features for both positive and negative texts. \rev{Directing the network to focus on the common information between the positive and negative texts.}

{\textit{Anchor}} is especially useful when we have two texts that differ from each other only by one word. Such is the case with our caption tree generation method and the VL-CheckList~\cite{vlc} dataset.
As we show in \cref{chapter:exp}, this leads to better interpretability and also to some performance gains as it makes the model more focused on the relevant parts of the image.

\subsection{ \oursifull{} (\oursi{}) }
Another way to get a single relevancy map when we have two input texts is to generate one relevancy map per text and then generate a new relevancy map by subtracting the negative relevancy map from the positive one. Although this method is not ``fair" as we use the knowledge of which caption is positive and which is negative, we can still leverage this method combined with {\textit{Token Removal}} for interpretability. The reason for that is that {\textit{\oursi{}}} provides importance scores to the image tokens and it does not `directly intervene' in the actual decision of the \vlm{}. Therefore, if we find that the tokens of high relevancy according to {\textit{\oursi{}}} correlate with better accuracy of the \vlm{}, then we can get a better understanding of which tokens (that correspond to locations at the input image) affect the decision of the \vlm{} and thus attain improved interpretability.  In Figure~\ref{fig:Anchor_DiRE_interpretability_graph} we show that indeed tokens of high relevancy according to {\textit{\oursi{}}} correlate with better accuracy of the \vlm{} (see details in Section~\ref{sec:Quantitative}).


\begin{figure}[t]
\centering
\begin{adjustbox}{max width=\columnwidth}
\includegraphics{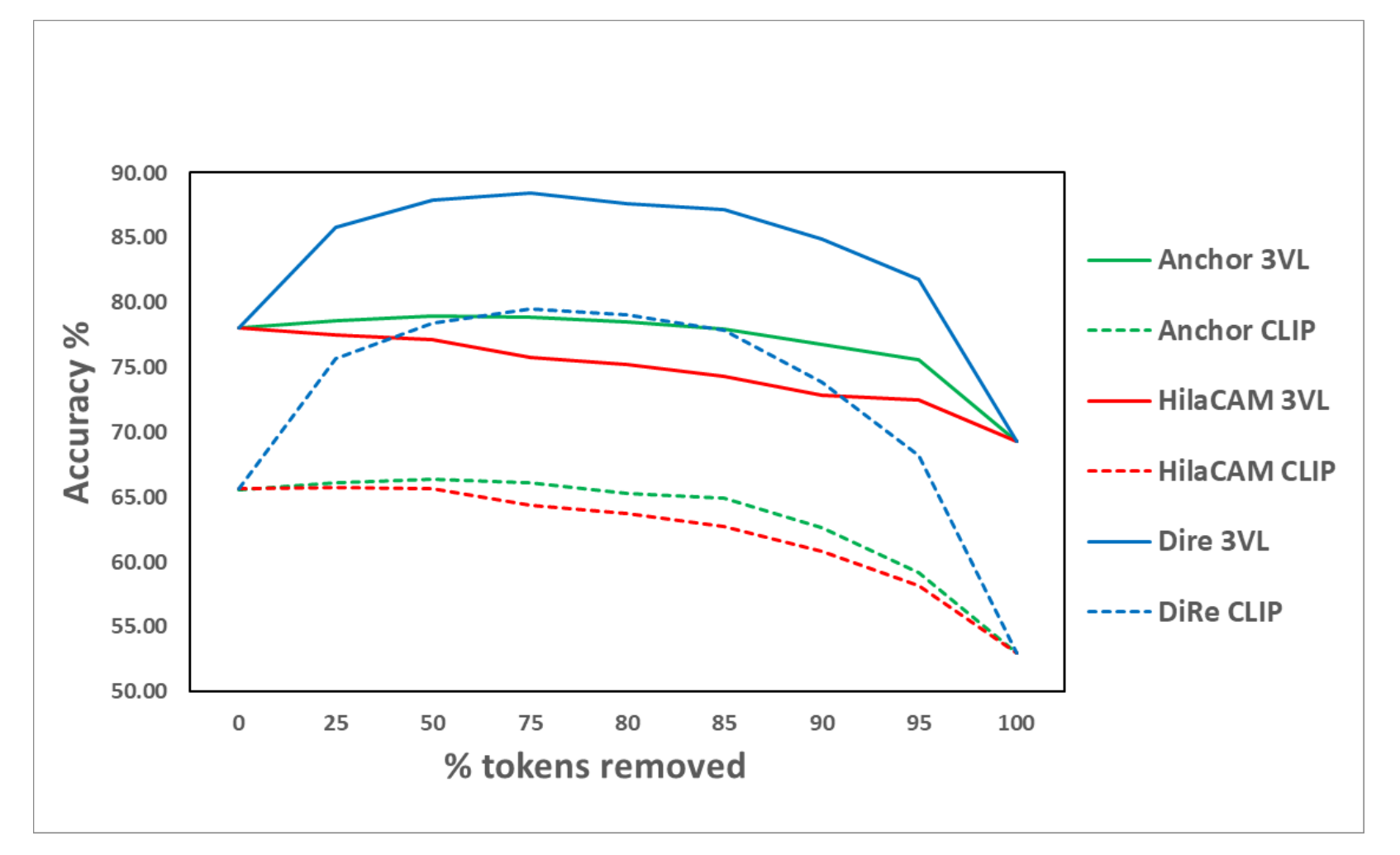}
\end{adjustbox}
\caption{3VL Token Removal accuracy on VL-Checklist (average of Attribute and Relation). HilaCAM vs. Anchor vs. {\textit{\oursi{}}} for both \oursm{} and vanilla CLIP. Notice that 3VL gets better accuracy compared to vanilla CLIP and its relative improvement with Anchor is better. Note that better improvement by {\textit{Token Removal}} indicates that a better understanding of the token importance is gained and therefore there is a better interpretability.}
\label{fig:Anchor_DiRE_interpretability_graph}
\end{figure}

\begin{table*}[t]
    \caption{Top-1 accuracy on VL-Checklist VG Relation and Attribute datasets, with openAI CLIP VIT-B/32, NegCLIP \cite{yuksekgonul2023when}, RB+LLM Negs \cite{doveh2023teaching}, and {\textit{\oursm{}}}. {TR} stands for {\textit{Token Removal}} (we remove from low to high importance). Token importance is taken from relevancy maps generated by \cite{HilaCAM} on the image with different texts. ANC = Anchor (ours) - takes the part of texts that differ from each other.}
    \label{tbl:VL-Check_A+R}
    \centering
    \begin{adjustbox}{width=0.8\textwidth}
    \begin{tabular}{|l|c|c|c|c|c|c|c|c|}
        \toprule
        Model & Att Color & Att Material & Att Size & Att Action & Att State & Rel Action & Rel Spatial & Avg \\ 
        \midrule
        CLIP \cite{clip} & 66.68 & 67.25 & 67.90 & 73.31 & 71.09 & 60.80 & 55.90 & 66.13 \\
        NegCLIP \cite{yuksekgonul2023when} & 73.62 & 75.83 & 66.12 & 75.37 & 70.24 & 66.18 & 60.87 & 69.75 \\
        RB+LLM Negs \cite{doveh2023teaching} & \textbf{82.70} & \textbf{84.90} & \textbf{78.10} & 71.60 & 75.13 & 70.00 & 78.40 & 77.26 \\
        \midrule
        Our 3VL & 75.29 & 82.63 & 70.90 & \textbf{81.10} & 75.03 & 81.69 & 81.14 & 78.25 \\
        Our 3VL + 65\% TR ANC & 76.18 & 82.51 & 71.36 & 79.57 & \textbf{76.63} & \textbf{83.73} & \textbf{84.29} & \textbf{79.18} \\
        \bottomrule
    \end{tabular}
    \end{adjustbox}
\end{table*}

\begin{table*}[t]
    \caption{Top-1 accuracy on VL-Checklist VG Object dataset, with openAI CLIP VIT-B/32, NegCLIP \cite{yuksekgonul2023when}, RB+LLM Negs \cite{doveh2023teaching}, and {\textit{\oursm{}}}. As an ablation, we also compare to CLIP and CLIP+LoRA \rev{which} were trained without tree-based loss.}
    \label{tbl:VL-Check Obj}
    \centering
    \begin{adjustbox}{width=0.8\textwidth}
    \begin{tabular}{|l|c|c|c|c|c|c|c|}
        \toprule
        Model & Location Center & Location Margin & Location Mid & Size Large & Size Medium & Size Small & Avg \\
        \midrule
        CLIP & 86.95 & 77.75 & 72.75 & 85.50 & 80.50 & 70.60 & 79.01 \\
        CLIP contrastive & 88.53 & 73.42 & 81.44 & 89.71 & 80.03 & 73.98 & 81.18 \\
        CLIP+LoRA contrastive & 88.04 & 72.29 & 80.77 & 89.62 & 79.35 & 71.68 & 80.29 \\
        NegCLIP & 89.61 & 72.64 & 81.88 & 90.33 & 80.82 & 73.95 & 81.54 \\
        RB+LLM Negs & 91.70 & 83.20 & 78.90 & 90.30 & 84.55 & 77.00 & 84.34 \\
        \midrule
        Our \textbf{\textit{\oursm{}}} & \bf{93.82} & \bf{85.02} & \bf{89.48} & \bf{94.96} & \bf{88.61} & \bf{83.78} & \bf{89.28} \\      
        \bottomrule
    \end{tabular}
    \end{adjustbox}
\vspace{-0.15in}
\end{table*}

\begin{table}[t]
    \caption{Top-1 accuracy on VL-Checklist(VLC) Relation(Rel) average, Attribute(Attr) average, Object(Obj) average, and average zero-shot(ZS) on ELEVATER 21 datasets. We test on openAI CLIP VIT-B/32, NegCLIP \cite{yuksekgonul2023when}, RB+LLM Negs \cite{doveh2023teaching}, and {\textit{\oursm{}}}. As an ablation, we also compare to CLIP and CLIP+LoRA that were trained without tree-based loss.}
          \vspace{-0.08in}
    \label{tbl:VLC_Elevater}
    \centering
    \begin{adjustbox}{max width=\columnwidth}
    \begin{tabular}{|l |c |c |c |c |c |}
        \toprule
        Model & VLC-Attr & VLC-Rel & VLC-Obj & VLC-AVG & AVG 21 ZS-Tasks \\
        \midrule
        CLIP \cite{clip} & 69.25 & 58.35 & 79.01 & 68.87 & 56.37 \\
        CLIP contrastive & 70.57 & 57.12 & 82.621 & 70.10 & 55.03 \\
        CLIP+LoRA contrastive & 69.46 & 55.27 & 80.29 & 68.34 & \textbf{56.47} \\
        NegCLIP \cite{yuksekgonul2023when} & 72.24 & 63.53 & 81.54 & 72.43 & 51.67\\
        RB+LLM Negs \cite{doveh2023teaching} & \textbf{78.49} & 74.20 & 84.34 & 79.01 & 54.66 \\
        \midrule
        Our 3VL CLIP & 76.99 & \textbf{81.42} & \textbf{89.28} & \textbf{82.46} & 52.56 \\
    \bottomrule
    \end{tabular}
    \end{adjustbox}
    \vspace{-0.08in}
\end{table}

\begin{table}[t]
    \caption{Top-1 accuracy on VSR \cite{Liu2022VisualSR} benchmark, with openAI CLIP VIT-B/32, NegCLIP \cite{yuksekgonul2023when}, RB+LLM Negs \cite{doveh2023teaching}, and {\textit{\oursm{}}}.}
    \label{tbl:VSR}
    \centering
    \begin{adjustbox}{max width=0.5\textwidth}
    \begin{tabular}{|l|c|}
        \toprule
        Model & Accuracy  \\ 
        \midrule
        CLIP \cite{clip} & 50.07 \\
        NegCLIP \cite{yuksekgonul2023when} & 49.33 \\
        RB+LLM Negs \cite{doveh2023teaching} & 50.21 \\
        \midrule
        Our 3VL & \textbf{51.98} \\
        \bottomrule
    \end{tabular}
    \end{adjustbox}
\end{table}

\begin{table}[t]
    \caption{COCO Image-Text retrieval with CLIP VIT-B/32, and \textbf{\textit{\oursm{}}}}
    \label{tbl:COCO_ret}
    \vspace{-0.08in}
    \centering
    \begin{adjustbox}{width=0.8\columnwidth}
    \begin{tabular}{c|c c c|c c c}
        \toprule
        \multirow[c]{2}{*}[0in]{} & \multicolumn{3}{c|}{I2T}  & \multicolumn{3}{c}{T2I}\\
        & R@1 & R@5 & R@10 & R@1 & R@5 & R@10\\
        \midrule       
        CLIP \cite{clip} & 32.54 & 57.7 & 68.08 & 28.66 & 53.04 & 64.44\\
        \midrule
        3VL & \textbf{33.72} & \textbf{62.08} & \textbf{73.12} & \textbf{36.54} & \textbf{63.32} & \textbf{74.48}\\
        \bottomrule
    \end{tabular}
    \end{adjustbox}
\end{table}

\begin{table}[t]
    \caption{FLICKR Image-Text retrieval with CLIP VIT-B/32 and \textbf{\textit{\oursm{}}}}
    \label{tbl:FLICKR_ret}
    \vspace{-0.08in}
    \centering
    \begin{adjustbox}{width=0.8\columnwidth}
    \begin{tabular}{c|c c c|c c c}
        \toprule
        \multirow[c]{2}{*}[0in]{} & \multicolumn{3}{c|}{I2T}  & \multicolumn{3}{c}{T2I}\\
        & R@1 & R@5 & R@10 & R@1 & R@5 & R@10 \\
        \midrule       
        CLIP \cite{clip} & 69.5 & 90.1 & 95 & 67 & 89.5 & 93.9\\
        \midrule
        3VL & \textbf{71} & \textbf{91.1} & \textbf{95.3} & \textbf{74} & \textbf{93.5} & \textbf{96.1}\\
        \bottomrule
    \end{tabular}
    \end{adjustbox}
    \vspace{-0.08in}
\end{table}

\begin{table}[t]
    \caption{CLIPSeg segmentation with CLIP VIT-B/32, and \textbf{\textit{\oursm{}}}}
    \label{tbl:CLIPSeg}
    \vspace{-0.08in}
    \centering
    \begin{adjustbox}{width=0.6\columnwidth}
    \begin{tabular}{c|c c |c c }
        \toprule
        \multirow[c]{2}{*}[0in]{} & \multicolumn{2}{c|}{PhraseCut}  & \multicolumn{2}{c}{COCO}\\
        & IoU\textsubscript{FG} & IoU\textsubscript{BIN} & IoU\textsubscript{FG} & IoU\textsubscript{BIN} \\
        \midrule       
        CLIP \cite{clip} & 52.4 & 71.7 & 54.8 & 73.2\\
        \midrule
        3VL & \textbf{53.4} & \textbf{72.1} & \textbf{57.2} & \textbf{74.7}\\
        \bottomrule
    \end{tabular}
    \end{adjustbox}
\end{table}

\section{Experiments}
\label{chapter:exp}

We evaluate our approach on popular \clcfull{} benchmarks and on general downstream tasks. Then, we perform an interpretability assessment and examine CLIP failure modes and biases. We conclude this section with an ablation study.


\subsection{\textit{\oursm{}} \clcfull{} Evaluation}
We start by evaluating \oursm{} on the VL-Checklist~\cite{vlc} benchmark. 

We evaluate {\textit{\oursm{}}} performance with and without {\textit{Token Removal}}. 
Table~\ref{tbl:VL-Check_A+R} presents the results on the VL-Checklist VG datasets \cite{vlc}, namely, VG Relation, Object, and Attribute. Note that the VG datasets are the most challenging ones in the VL-Checklist benchmark.
In the experiments, we compare our approach to openAI VIT-B/32 CLIP \cite{clip}, NegCLIP \cite{yuksekgonul2023when}, RB+LLM Negs \cite{doveh2023teaching}.
The evaluation of the models' performance involved measuring the Top-1 accuracy metric. Note the improvement in 3VL compared to the CLIP baseline and that our approach improves over the other techniques in most cases. 

In addition, we evaluated the impact of using {\textit{Token Removal}} {(TR)}, where tokens were systematically removed from low to high importance according to {\textit{Anchor}} (ANC) relevancy map as explained above.
Note that {\textit{Token Removal}} {(TR)} together with {\textit{Anchor}} improves {\textit{\oursm{}}} performance.

In Table \ref{tbl:VL-Check Obj} we add detailed results of VL-Checklist Object datasets, where we evaluate objects' location invariance(e.g, center, middle, and margin) and size invariance (e.g., small, large, medium). A robust VLM should recognize objects' existence, regardless of their location and size. Note the substantial performance improvement over all Object categories.

Table~\ref{tbl:VLC_Elevater} summarizes the results for the group categories of VL-Checklist. Notice that {\textit{\oursm{}}} is significantly better than the other techniques on the relational and object categories. This may be attributed to the tree structure we introduce during training that helps in such type of queries. Yet, in attributes, the improvement is better than NegCLIP but less than \cite{doveh2023teaching}. We conjecture that creating shallow rule-based hard negatives is more efficient for teaching attributes concepts like color and material while our tree structure is more beneficial for complex relational knowledge.

In addition, we report the results of 21 zero-shot tasks (ZS-Tasks). This includes Imagenet and the 20 image classification datasets of ELEVATER \cite{elevater}. This test checks the zero-shot image classification performance of our model and it shows that there is only little forgetting in our learning process. \textit{\oursm{}} is better in this respect than NegCLIP but slightly worse than \cite{doveh2023teaching}.
We also compare to CLIP and CLIP+LoRA finetuned on MS-COCO with contrastive loss (without tree-augmented loss) as an ablation of the effects of finetuning on MSCOCO.

\subsection{\textit{\oursm{}} Visual Spatial Reasoning (VSR) Evaluation}
The VSR \cite{Liu2022VisualSR} benchmark evaluates \vlm{}s' spatial reasoning using image-to-text retrieval. Each image contains a visible spatial relation, and for each image, we have a pair of captions with a description of this visible spatial relation and textual ``True" and ``False" labels concatenated at the end (e.g. ``A is in front of B (False)", ``A is in front of B (True)"). The model needs to retrieve the caption with the correct label.

Table~\ref{tbl:VSR} presents the results of the VSR \cite{Liu2022VisualSR} dataset with openAI CLIP VIT-B/32, NegCLIP \cite{yuksekgonul2023when}, RB+LLM Negs \cite{doveh2023teaching}, and {\textit{\oursm{}}}.
Although this task is very hard for all models (probably somewhat because of the unusual caption + label prompt style) we can still see an advantage for {\textit{\oursm{}}}.

\subsection{\textit{\oursm{}} Downstream Tasks Evaluation}

We experimented on downstream tasks with the original CLIP model replaced by our {\textit{\oursm{}}} and show the performance improvement of {\textit{\oursm{}}} over CLIP.


\noindent \textbf{Image-Text retrieval.}
In Tables \ref{tbl:COCO_ret} and \ref{tbl:FLICKR_ret} we report results of image to text (I2T) and text to image (T2I) retrieval on COCO and FLICKR datasets. 

\noindent \textbf{Image Segmentation.} Table \ref{tbl:CLIPSeg} reports segmentation results using CLIPSeg \cite{lueddecke22_cvpr} with vanilla CLIP and with {\textit{\oursm{}}} on COCO and PhraseCut \cite{wu2020phrasecut} datasets.



\begin{figure*}[t]
\centering
\begin{adjustbox}{width=\textwidth}
\includegraphics{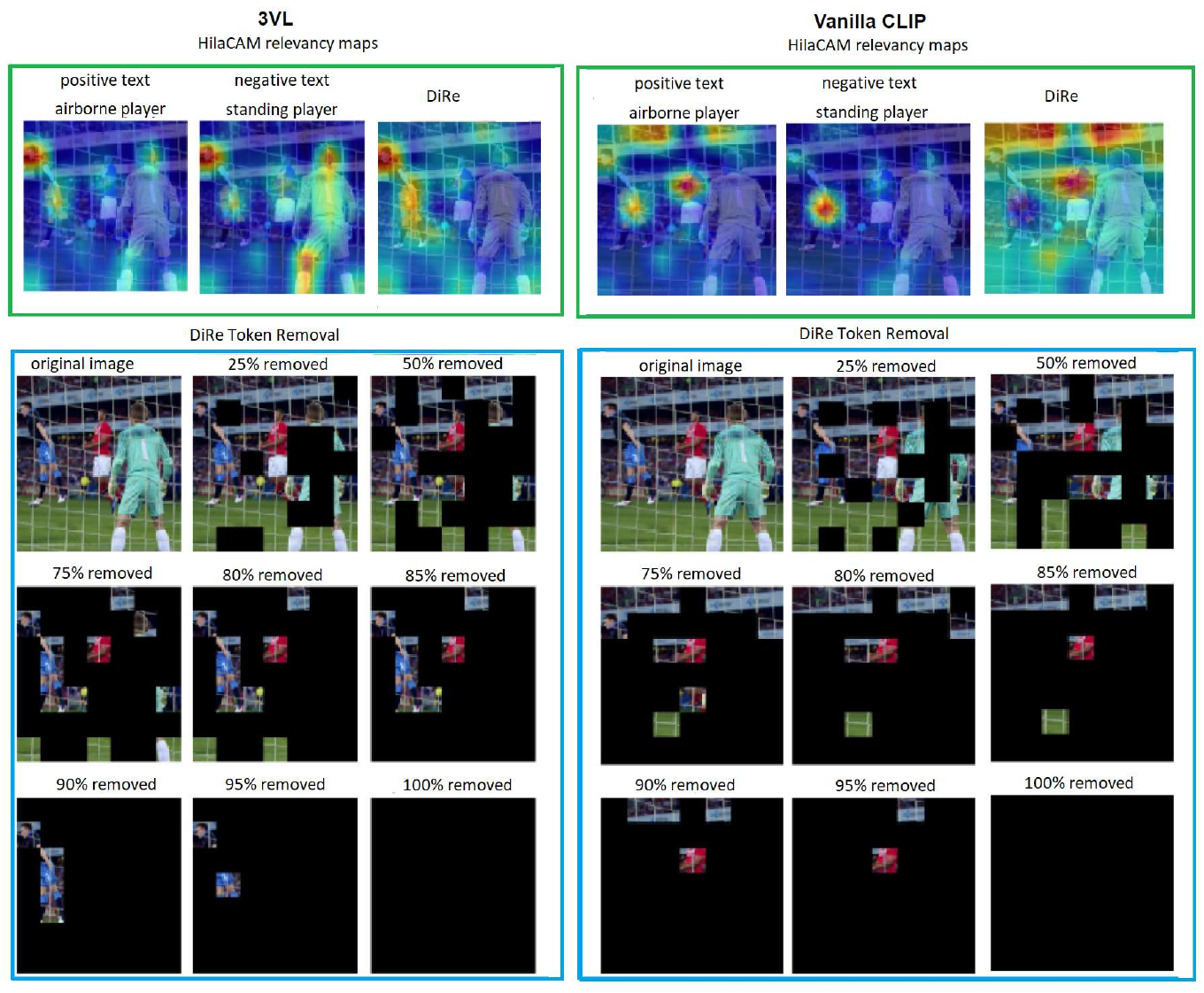}
\end{adjustbox}
\vspace{-0.6in}
\caption{A comparison of vanilla CLIP (right) and {\textit{\oursm{}}} (left) interpretability for a randomly selected sample from the VL-Checklist VG attribute action dataset, where both models are mistaken.  
At the top we present the HilaCAM \cite{HilaCAM} relevancy map visualizations (red areas important, purple not important) for the positive text (``airborne player") and negative text (``standing player") and the {\textit{\oursi{}}} relevancy map. On the second row from left to right, we can see the {\textit{Token Removal}} process. Starting from the original image we gradually remove the least significant tokens according to the {\textit{\oursi{}}} relevancy map.
}
\label{fig:VLC attr act}
\end{figure*}

\subsection{Interpretability Quantitative Assessment} \label{sec:Quantitative} 

\textbf{Assessing Interpretability via Token Removal.}
To assess the interpretability of our method, we follow the negative perturbation ({\textit{Token Removal}}) test in \cite{HilaCAM}. First, a relevancy map is extracted from the image-text pair. Then, we remove image tokens in order of increasing importance (from lowest to highest importance according to the relevancy map) and use the image with removed tokens as input to the image encoder and calculate the cosine similarity to both the positive text and negative text. As we have two texts, positive and negative, for HilaCAM \cite{HilaCAM} we calculate one relevancy map for the positive text and one for the negative text. Then, we remove tokens according to the relevancy map created by each of the texts and get two subsets of tokens which we calculate cosine similarity for. For {\textit{Anchor}} and {\textit{\oursi{}}} we have only one relevancy map and we use it for {\textit{Token Removal}}. The prediction in all cases is the text with the highest similarity score to the input image after {\textit{Token Removal}}.

Figure \ref{fig:Anchor_DiRE_interpretability_graph} presents a negative perturbation test of 3VL and vanilla CLIP models together with HilaCAM \cite{HilaCAM}, {\textit{Anchor}} and {\textit{\oursi{}}} relevancy maps.
A more gradual decline in performance should point to a more explainable method as we remove less important tokens where importance is taken from that method's relevancy map. We can see that {\textit{Anchor}} {\textit{Token Removal}} can improve performance which suggests that removing tokens using {\textit{Anchor}} relevancy map focuses the model on the more important parts of the image. This strategy can also be used at inference time to improve the performance of \vlm{}s.


{\textit{\oursi{}}} creates a new relevancy map by subtracting the negative relevancy map from the positive relevancy map (both obtained from HilaCAM). Note that while {\textit{\oursi{}}} cannot be used to improve accuracy at inference time (since it uses ground truth information about which input is positive and which is negative), it can be used to improve the interpretability of \vlm{}s when analyzing a model and the ground truth is available.
Figure~\ref{fig:Anchor_DiRE_interpretability_graph} shows the advantage of {\textit{\oursi{}}} for interpretability purposes as it emphasizes better the most informative tokens in the image.

The Token Removal evaluation shows that removal of less relevant tokens by our methods “Anchor” and DiRe” leads to greater performance improvement than token removal using other methods’ relevancy maps. Thus, we find that our generated relevancy maps are more “accurate” than others.

\textbf{Assessing Interpretability via User Study.}
To further assess the interpretability of {\textit{\oursm{}}} we conducted a user study (for which we got the necessary IRB Approval). 
In the user study we have compared {\textit{\oursm{}}} to vanilla CLIP using relevancy maps generated from image-text pairs with {\textit{HilaCAM}}, and we have compared relevancy maps generated by {\textit{HilaCAM}} to relevancy maps generated by {\textit{Anchor}} and {\textit{\oursi{}}}. We have compared the different relevancy maps by letting users decide which relevancy map emphasizes a caption in an image better. We randomly sampled 25 images from the VL-Checklist VG dataset. For each image we presented the original image on the right and the relevancy maps (either {\textit{HilaCAM}} relevancy maps for {\textit{\oursm{}}} and vanilla CLIP, or the relevancy maps of {\textit{HilaCAM}}, {\textit{Anchor}} and {\textit{\oursi{}}}) in random order.  
A total of 37 respondents participated in the study and were asked ``which image better emphasizes the caption..." (a), (b), (c) or neither. We have found {\textit{\oursm{}}} maps to be more explainable than vanilla CLIP. 51.67\% of user choices were {\textit{\oursm{}}}, 27.4\% vanilla CLIP and 20.93\% neither. Moreover, in 78.94\% of the questions {\textit{\oursm{}}} got the most votes. We have also found {\textit{\oursi{}}} to be more explainable than {\textit{Anchor}}, which is more explainable than {\textit{HilaCAM}}. 57.6\% of user choices were {\textit{\oursi{}}}, 19.2\% {\textit{Anchor}}, 11.2\% {\textit{HilaCAM}} and 12\% neither. Moreover, in 83.33\% of the questions {\textit{\oursi{}}} got the most votes and in 16.67\% of the questions {\textit{Anchor}} got the most votes. {\textit{HilaCAM}} did not get the most votes in any of the questions. Some screenshots from the user study appear in the appendix in Figure ~\ref{fig:study}.

\rev{The user study shows that users have a preference for our relevancy maps. This may imply that our relevancy maps align more closely with human interpretation by highlighting important image parts. However, it is possible that while our relevancy maps are more visually appealing to users, they do not necessarily explain the model's decision better. We leave a further exploration of the benefit of our relevancy map approach to a future research.}

\begin{figure*}[t]
\vspace{-0.5cm}
\centering
\begin{adjustbox}{width=\textwidth}
\includegraphics{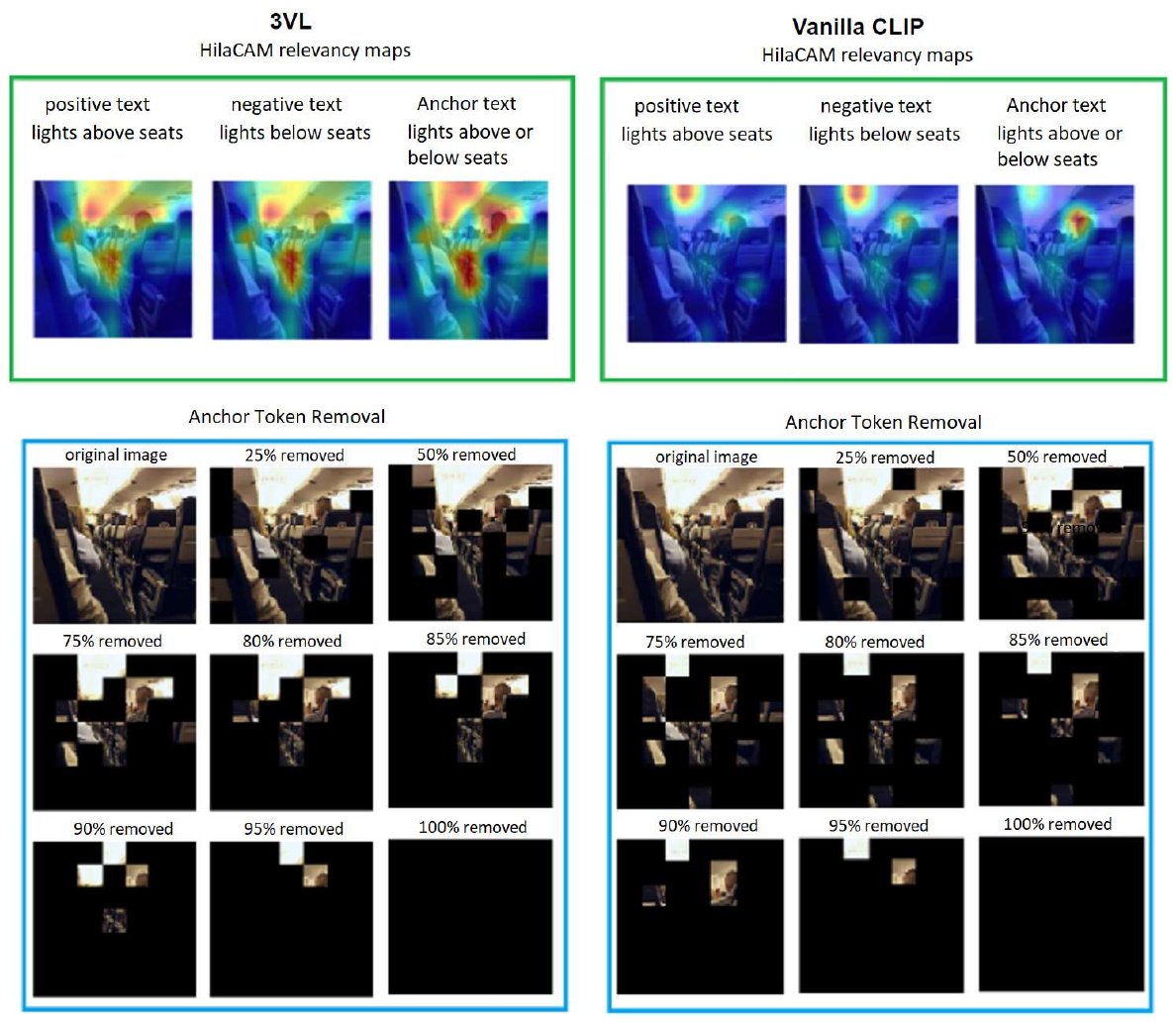}
\end{adjustbox}
\vspace{-0.8in}
\caption{A comparison between \oursm{}(left) and vanilla CLIP(right) visualizations with Anchor token removal. At the top: HilaCAM relevancy maps for positive text, negative text, and Anchor text ("light above seats", "light below seats", "light above or below seats"). At the bottom the input image with token removal (from 0\% removal to 100\%) according to Anchor relevancy map. \oursm{} attends to all lights in the photo, whereas Vanilla CLIP attends more to lights above seats. After Token removal, both attend to lights above seats.}
\label{fig:VLC_vis6}
\vspace{-0.5cm}
\end{figure*}

\begin{figure}[t]
\centering
\begin{adjustbox}{max width=\columnwidth}
\includegraphics[scale=1]{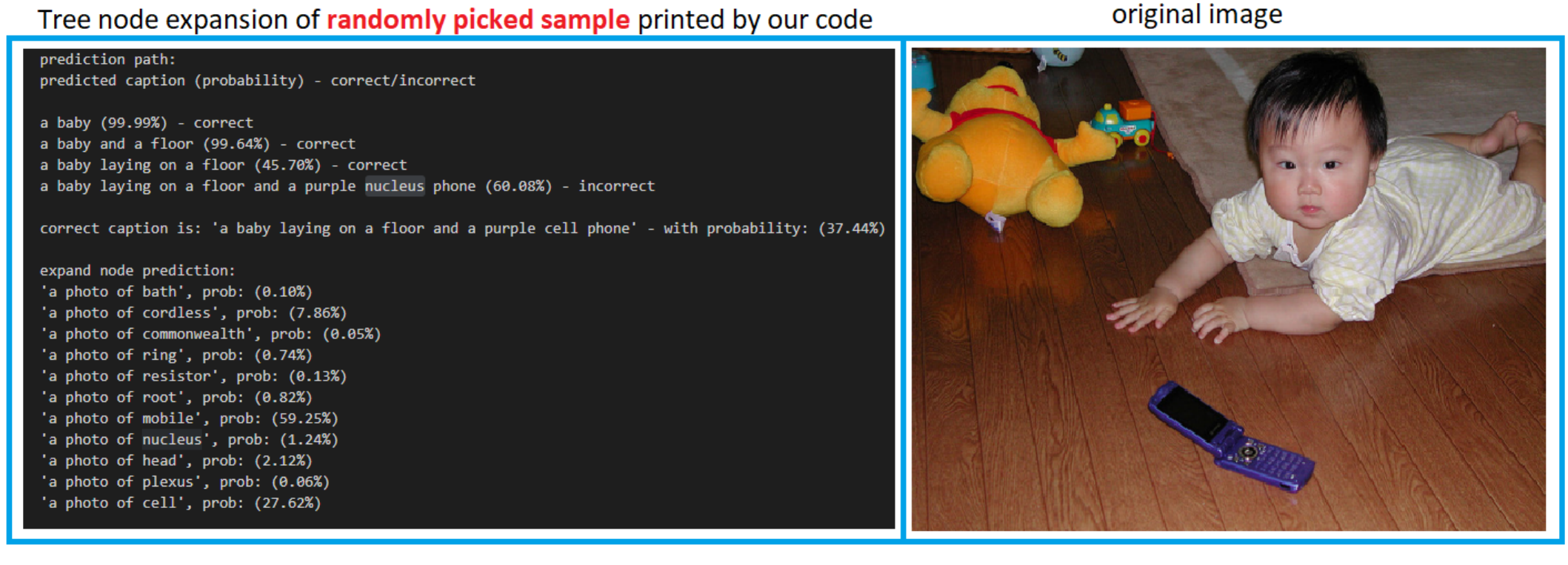}
\end{adjustbox}
\vspace{-0.25in}
\caption{Tree node expansion when the model confuses 'nucleus phone' and 'cell phone'. The model fails to understand the term 'cell phone' and interprets the word 'cell' as part of an organism, but when given the word 'mobile' instead of 'cell', the model identifies the mobile phone in the image and gives it a much higher probability than 'nucleus'.}
\label{fig:cell}
\vspace{-0.05in}
\end{figure}

\begin{figure}[t]
\centering
\begin{adjustbox}{max width=\columnwidth}
\includegraphics[scale=1]{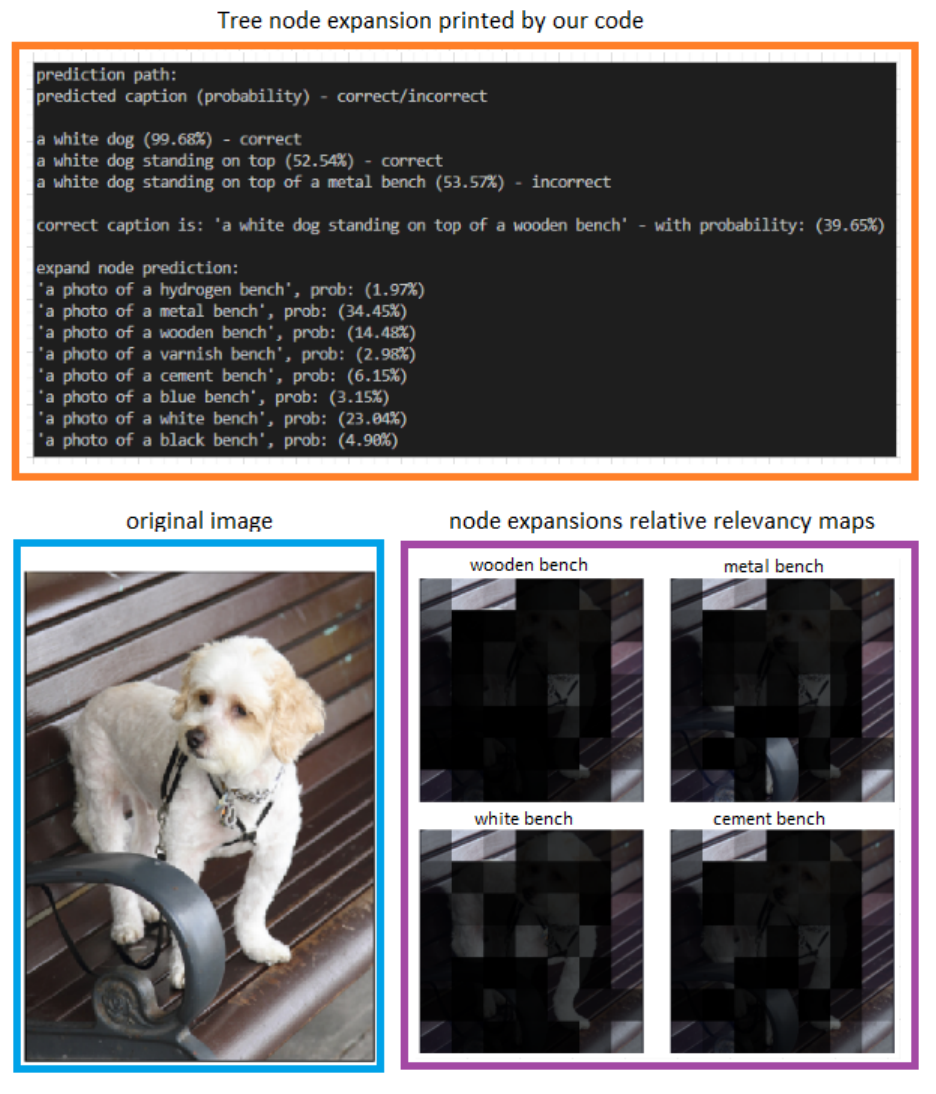}
\end{adjustbox}
\vspace{-0.28in}
\caption{A simple example of tree node expansion printed by our code (top left) in case of failure, the original image (bottom left), and 
relative relevancy maps (bottom right) of the top 4 most probable captions.}
\label{fig:COCO}
\vspace{-0.1in}
\end{figure}

\subsection{Interpretability Qualitative assessment} 
Figure~\ref{fig:VLC attr act} visualizes the strong interpretability of {\textit{\oursm{}}} with HilaCAM and {\textit{\oursi{}}}. We compare the visualizations of vanilla CLIP and {\textit{\oursm{}}} with HilaCAM \cite{HilaCAM} and \textit{\oursm{}} on a random sample from the VL-Checklist dataset, where both CLIP and {\textit{\oursm{}}} are mistaken.
Note that from the HilaCAM visualizations, it is hard to understand the source of the error. With {\textit{\oursm{}}} the visualization of the positive text emphasizes an airborne player and the visualization of the negative text emphasizes both an airborne player and a standing player. With Vanilla CLIP the positive and negative visualizations emphasize different parts of airborne players. Yet both models mistakenly predict ``standing player". When adding {\textit{\oursi{}}} visualization we understand that the model should have focused more on the bottom part of the airborne player in addition to the upper part to understand that he is airborne. The {\textit{Token Removal}} emphasizes this further. Figure~\ref{fig:VLC_vis6} visualizes the strong interpretability of {\textit{\oursm{}}} with HilaCAM and {\textit{Anchor}. \oursm{} attends to all lights in the photo, whereas Vanilla CLIP attends more to lights above seats. After Token removal, both attend to lights above seats.

\subsection{Using {\textit{\oursm{}}} for Understanding VLMs' Failures}
\label{sec:failure_cases}

Using our caption tree generation method we are able to find exact words and part of speech tags that the model fails to recognize. To offer extra insight regarding \rev{the} model's decisions, we expand the caption tree at the level in which the model failed, with extra words generated to be similar (co-hyponyms and synonyms) to the chosen negative word and to the positive word and check \rev{the} probability for all these words. We also add relative relevancy visualizations for these words at failure points. 
Figure \ref{fig:cell} demonstrates how CLIP fails to understand \rev{multiple-word phrases} and considers each single word on its own. Figure \ref{fig:COCO} shows the relative relevancy maps of the 4 most probable captions according to CLIP, highlighting different parts of the image for each(e.g. the wooden bench and the metal bench rail).  

\begin{table}[t]
    \caption{Number of failures on COCO testset per part of speech tag. In addition to the absolute number of failures, for NegCLIP and \textit{\oursm{}} we present the \% improvement over CLIP. For vanilla CLIP, NegCLIP \cite{yuksekgonul2023when} and {\textit{\oursm{}}}. {\textit{\oursm{}}} improves by more than 50\% on verbs(VERB) and by 46\% on adpositions(ADP).}
    \label{tbl:POS}
    \vspace{-0.08in}
    \centering
    \begin{adjustbox}{width=\columnwidth}
    \begin{tabular}{|l|c|c|c|c|}
        \toprule
        Model & NOUN & ADP & VERB & ADJ \\
        \midrule
        CLIP & 3104 & 2927 & 1612 & 913 \\
        NegCLIP & 2743 / 11.63\% & 1989 / 32.04\% & 900 / 44.16\% & 695 / 23.87\% \\
        \midrule
        3VL & 1976 / 36.34\%
 & 1575 / 46.19\% & 780 / 51.61\% & 612 / 32.96\% \\
        \bottomrule
    \end{tabular}
    \end{adjustbox}
    \vspace{-0.08in}
\end{table}

In Table \ref{tbl:POS} we show statistics of failures per part-of-speech tag on \rev{the} MSCOCO test set with pretrained CLIP, ARO NegCLIP \cite{yuksekgonul2023when} and {\textit{\oursm{}}}. Note the big improvement in some of CLIP's biggest weaknesses, verbs and adpositions understanding, with more than 50\% improvement on verbs and 46\% improvement on adpositions.

To gain a better understanding of the model's success and failures, we analyze the more frequent word failures and find some strong biases in vanilla CLIP. We show in Table \ref{tbl:bias} some CLIP biases and how it change with {\textit{\oursm{}}}. 
Note that removing less relevant tokens may enhance already existing biases. Yet, Table \ref{tbl:bias} shows that this does not amplify the bias. Despite that it is clear that further reduction of the model bias is still required and our strategy just slighly improves it.

\subsection{Ablation Study}
We conduct an ablation study to assess the impact of different factors on the effectiveness of our proposed tree-based training technique. We examine the effects of different loss functions, different tree structures, and negatives generation. We examine the regularization effects of limiting our caption tree to a maximum depth or limiting the number of maximum negatives generated in each tree.

\noindent {\textbf{Different loss functions ablation.}}
In Table~\ref{tbl:VLC_Elevater} we show the impact of different loss functions on the performance of our model. Note that the addition of LoRA degrades a bit the performance on VL-checklist but reduces the forgetting effect (exhibited in the performance on ELEVATER~\cite{elevater}). Note that the addition of tree loss leads to a significant improvement over just using contrastive learning. 

\begin{table}[t]
    \caption{CLIP fail rate on some pairs of positive-negative words on COCO test set. }
    \label{tbl:bias}
    \vspace{-0.08in}
    \centering
    \begin{adjustbox}{max width=0.9\columnwidth}
    \begin{tabular}{|c|c|c|c|}
        \toprule
        positive word & negative word & CLIP fail rate (\%) & 3VL fail rate (\%) \\
        \midrule
        small & large & 50 & 35.8 \\
        large & small & 25.26 & 22.1 \\
        on & off & 35.74 & 20.5 \\
        off & on & 78.26 & 50 \\
        young & old & 23.57 & 4.5 \\
        old & young & 50 & 28.2 \\
        down & up & 55.26 & 35.03 \\
        up & down & 44.08 & 19.2 \\
        white & black & 22.17 & 13.87 \\
        black & white & 27.35 & 23.36 \\
        man & woman & 13.77 & 6.83 \\
        woman & man & 17.6 & 14.18 \\
        \bottomrule
    \end{tabular}
    \end{adjustbox}
    \vspace{-0.15in}
\end{table}

\noindent \textbf{Caption tree variants ablation.}
In the tree training process of \textit{\oursm{}}, we have explored many variants of caption tree structure, different negatives generation methods and their combinations. We provide below a detailed comparison between the different options using the example caption {\textit{``several people standing in a green field together while flying kites"}}.

\begin{itemize}[leftmargin=*]

\item \emph{Tree structures.} We explored two main tree structures:
\begin{enumerate}
    
\item basic tree structure ({\textit{``Basic"}}) where each noun phrase appears alone (without adding previous noun phrases and connecting text) in different tree levels and the full caption in the last tree level. (e.g. for the above example we will get a tree with 3 levels. {\textit{``several people"}} in the first level, {\textit{``a green field"}} in the second level, and {\textit{``several people standing in a green field together while flying kites"}} in the third level). 
\item incremental tree structure ({\textit{``Incremental"}}) where each noun phrase is prepended once with previous tree level text only and once with previous tree level text plus the original text that connects to the current noun phrase (verbs and adpositions). (e.g. for the above example we will get a tree with 4 levels. {\textit{``several people"}} in the first level, {\textit{``several people and a green field"}} in the second level, {\textit{``several people standing in a green field"}} in the third level, and {\textit{``several people standing in a green field together while flying kites"}} in the fourth level).

\end{enumerate}

\item \emph{Negatives generation.} We explored three main methods:

\begin{enumerate}

\item Using WordNet ({\textit{``WN"}}) replace nouns and verbs with co-hyponyms and replace adjectives and adpositions with antonyms or with random adjectives or adpositions.

\item {\textit{``LLM prompt"}} - we generate co-hyponyms, antonyms, and random adjectives or adpositions with a prompt to FLAN-T5 (e.g., {\textit{``find an opposite for the word: $<>$"}}).

\item LLM mask completion ({\textit{``LLM mask"}}) - generate a negative word by replacing a positive word with a mask token and passing it to T5 LLM \cite{chung2022scaling} for mask completion. 
\textit{Other methods include also replacing only nouns or verbs with mask completion.}

\end{enumerate}

\item \emph{Negatives from intermediate tree levels.} We explored replacing also negatives from intermediate tree levels. For the example caption {\textit{``several people standing in a green field together while flying kites"}} we get the following tree levels:

\begin{enumerate}

\item {{\textit{``several people"}}, {\textit{``several animals"}}, {\textit{``one people"}}}

\item {{\textit{``several people and a green field"}}, {\textit{``several animals and a green field"}}, {\textit{``one people and a green field"}}, {\textit{``several people and a blue field"}}, {\textit{``several people and a green forest"}} }

\item {{\textit{``several people standing in a green field"}}, {\textit{``several animals standing in a green field"}}, {\textit{``one people standing in a green field"}}, {\textit{``several people standing in a blue field"}}, {\textit{``several people standing in a green forest"}}, {\textit{``several people standing out a green field"}}, {\textit{``several people sitting in a green field"}} }

\item {{\textit{``several people standing in a green field together while flying kites"}}, {\textit{``several animals standing in a green field together while flying kites"}}, {\textit{``one people standing in a green field together while flying kites"}}, {\textit{``several people standing in a blue field together while flying kites"}}, {\textit{``several people standing in a green forest together while flying kites"}}, {\textit{``several people standing out a green field together while flying kites"}}, {\textit{``several people sitting in a green field together while flying kites"}}, {\textit{``several people standing in a green field together while soaring kites"}}, {\textit{``several people standing in a green field together while flying sales"}} }

\end{enumerate}

\end{itemize}

In Table~\ref{tbl:trees} we show Top-1 accuracy results on VL-Checklist with six such tree variants. (i) {\textit{``Basic WN"}} - {\textit{``Basic"}} tree structure with WordNet negatives generation, (ii) {\textit{``Incremental WN"}} - {\textit{``Incremental"}} tree structure with WordNet negatives generation, (iii) {\textit{``WN+LLM prompt+mask"}} - {\textit{``Incremental"}} tree structure with {\textit{``LLM prompt"}} for opposites generation, T5 mask completion for words with no opposite and WordNet replacement if T5 mask completion generates a word with similar meaning.
(iv) {\textit{``WN+LLM prompt+inter negs"}} -  {\textit{``Incremental"}} tree structure with {\textit{``LLM prompt"}} for opposites generation and WordNet replacement if no opposite exists while using {\textit{Negatives from intermediate tree levels}}.
(v) {\textit{``WN+LLM prompt+mix inter negs"}} -  {\textit{``Incremental"}} tree structure with {\textit{``LLM prompt"}} for opposites generation and WordNet replacement if no opposite exists and mixing trees with {\textit{Negatives from intermediate tree levels}} and without.
(vi) {\textit{``WN+LLM prompt (\oursm{})"}} -  {\textit{``Incremental"}} tree structure with {\textit{``LLM prompt"}} for opposites generation and WordNet replacement if no opposite exists (without {\textit{Negatives from intermediate tree levels}}). 

For each such variant, we report the highest result attained by hyperparameter sweep on the validation set of MSCOCO. 
We found that LLM mask completion oftentimes generated positive caption alternatives instead of negative captions and we did not use it in the final model.
Best results were achieved using (vi) {\textit{``WN+LLM prompt"}} and therefore, we used it in the final {\textit{\oursm{}}} model.

\begin{table*}[t]
    \caption{Top-1 accuracy on VL-Checklist for different caption trees. (i) {``Basic WN"} - Basic tree structure with WordNet replacement, (ii) {``Incremental WN"} - Incremental tree structure with WordNet replacement, (iii) {``WN+LLM prompt+mask"} - Incremental tree, FLAN-T5 prompt for opposites, T5 mask completion if opposite doesn't exist and WordNet replacement if mask completion did not generate a word with a different meaning, (iv) {``WN+LLM prompt+inter negs"} - Incremental tree, FLAN-T5 prompt for opposites, and WordNet replacement if the opposite does not exist and using Negatives from intermediate tree levels, (v) {``WN+LLM prompt+mix inter negs"} - Incremental tree, FLAN-T5 prompt for opposites, and WordNet replacement if the opposite does not exist and mixing trees with Negatives from intermediate tree levels and without, (vi) {``WN+LLM prompt (\oursm{})"} - 
    Incremental tree, FLAN-T5 prompt for opposites, and WordNet replacement if the opposite does not exist (without Negatives from intermediate tree levels).}
          \vspace{-0.1in}
    \label{tbl:trees}
    \centering
    \begin{adjustbox}{max width=1.5\columnwidth}
    \begin{tabular}{|l|c|c|c|c|c|c|c|c|}
        \toprule
        Tree Method & Att color & Att material & Att size & Att action & Att state & Rel action & Rel spatial & Avg \\
        \midrule
        Basic WN & 70.54 & 75.79 & 69.72 & 75.95 & 70.46 & 81.81 & 73.94 & 74.03 \\
        Incremental WN & 74.41 & 79.23 & 70.13 & 78.12 & 71.63 & 83.81 & 75.70 & 76.15 \\
        WN+LLM prompt+mask & 72.31 & 77.53 & 69.98 & 76.69 & 69.39 & 83.43 & 76.87 & 75.17 \\
        WN+LLM prompt+inter negs & 73.91 & 77.81 & 68.94 & 76.11 & 71.94 & 80.59 & 78.18 & 75.35 \\
        WN+LLM prompt+mix inter negs & 73.79 & 78.10 & 69.05 & 76.64 & 72.58 & 80.67 & 78.24 & 5.58 \\
        \midrule
        WN+LLM prompt (3VL) & 75.29 & 82.63 & 70.90 & 81.10 & 75.03 & 81.69 & 81.14 & 78.25 \\
        \bottomrule
    \end{tabular}
    \end{adjustbox}
      \vspace{-0.1in}
\end{table*}

\subsection{{Maximum tree depth ablation}}
In Table~\ref{tbl:depth_ablation} we report Top-1 accuracy on the VL-Checklist when constraining the {\textit{3VL}} tree to maximum depth. We note that we don't limit here the number of generated negatives per caption, just the structure of the tree. So, for example, a tree of depth 1 will contain the same number of negatives as in the full caption tree, just in a single level. We can see that even though the number of negatives remains the same, the increase in tree depth contributes to improved performance. 

\begin{table}[t]
  \centering
  \caption{Top-1 accuracy on VL-Checklist when constraining {\textit{3VL}} tree to maximum depth}
        \vspace{-0.1in}
  \label{tbl:depth_ablation}
  \begin{adjustbox}{max width=\linewidth}
  \begin{tabular}{|l|l|l|l|l|l|l|l|l|}
    \hline
    Depth & Att color & Att material & Att size & Att action & Att state & Rel action & Rel spatial & Avg \\
    \hline
    1 & 72.70 & 78.42 & 69.28 & 78.81 & 70.99 & 78.54 & 74.10 & 74.69 \\
    2 & 72.99 & 79.51 & 69.24 & 79.81 & 72.90 & 79.04 & 74.49 & 75.43 \\
    3 & 73.27 & 79.43 & 68.68 & 80.97 & 72.79 & 80.99 & 78.55 & 76.38 \\
    \hline
  \end{tabular}
  \end{adjustbox}
\end{table}

\subsection{{Maximum negatives per tree ablation}}
In Table~\ref{tbl:max_negs_ablation} we report Top-1 accuracy on VL-Checklist when constraining the maximum total number of generated negatives per tree. We notice here the benefit in generating more negatives per caption.

\begin{table}[t]
  \centering
  \vspace{-0.1in}
  \caption{Top-1 accuracy on VL-Checklist when constraining the maximum total number of generated negatives per tree}
        \vspace{-0.1in}
  \label{tbl:max_negs_ablation}
  \begin{adjustbox}{max width=\linewidth}
  \begin{tabular}{|l|l|l|l|l|l|l|l|l|}
    \hline
    Max negatives & Att color & Att material & Att size & Att action & Att state & Rel action & Rel spatial & Avg \\
    \hline
    1 & 72.47 & 75.22 & 66.35 & 73.31 & 72.69 & 69.80 & 53.70 & 69.08 \\
    2 & 73.24 & 77.49 & 69.39 & 75.05 & 73.96 & 75.15 & 65.67 & 72.85 \\
    4 & 74.16 & 78.18 & 70.28 & 77.59 & 73.54 & 79.86 & 74.18 & 75.39 \\
    8 & 74.29 & 79.72 & 70.54 & 76.06 & 75.56 & 80.31 & 79.06 & 76.50 \\
    \hline
  \end{tabular}
  \end{adjustbox}
  \vspace{-0.08in}
\end{table}

\subsection{Limitations} 

Note that although {\textit{\oursm{}}} improves upon vanilla CLIP in all words, we sometimes see an increase in bias towards one word of a pair: Some pairs of words have a big gap between the fail rate when one of the words is the positive one and the other one is the negative word, compared to the fail rate with the same pair but with the positive and negative words switched. For example, as shown in Table \ref{tbl:bias}, when the word ``small" is positive and the word ``large" is negative CLIP has a 50\% fail rate, but when the word ``large" is positive and the word ``small" is negative the fail rate is just above 25\% which suggests a very strong bias of CLIP to the word ``large" over ``small". With {\textit{\oursm{}}} these fail rates are reduced to 35.8\% and 22.1\% and the gap is smaller. Yet, with the pairs ``on"-``off" and ``young"-``old" the bias is reduced a lot less with {\textit{\oursm{}}}. Moreover, for the pairs ``down"-``up",  ``black"-``white", and ``man"-``woman" the gap even increases, although the model makes fewer mistakes overall. This is because the COCO train set contains an inherent large bias to one word from each of these pairs. 

Having said that, we believe that the tree structure used with {\textit{\oursm{}}} can be employed to decrease this gap. Given that we have the tree structure, we encourage a greater penalty on the tree leaves for text that appears less in the training data. We believe that our proposed tree augmentation method can be utilized to reduce biases in \vlm{} training but we leave this direction to future work.

\section{Conclusion}
\label{chapter:sum}


In this work, we propose a novel approach to address the limitations of VLMs in \clcfull{} understanding and to enhance the interpretability of VLMs by leveraging the hierarchical structure of natural language. Our approach offers a richer exploration of the text space using several levels of incremental text augmentations from coarse to fine-grained, and in the process allows for better interpretation and analysis of model decisions. We complement our method with image heatmap relevancy maps of positive and negative texts and use this to identify further weaknesses of VLMs. We propose a new inference technique that effectively filters out the noise and irrelevant information from the input, allowing us to gain insights into the underlying failure modes of the model and achieve state-of-the-art performance on popular CLC benchmarks. We believe that our proposed tree augmentation method can be utilized to reduce biases in VLM training but we leave this direction to future work.
Our work has important implications for the development of more interpretable and effective VLMs for natural language understanding and computer vision.
Note that our proposed solution is generic. We have demonstrated it here with CLIP. Yet, future work can easily incorporate it in other \vlm{}s.

\bibliographystyle{IEEEtran}
\bibliography{references}

\begin{figure*}[t]
\centering
\begin{adjustbox}{max width=\textwidth}
\includegraphics[scale=0.7]{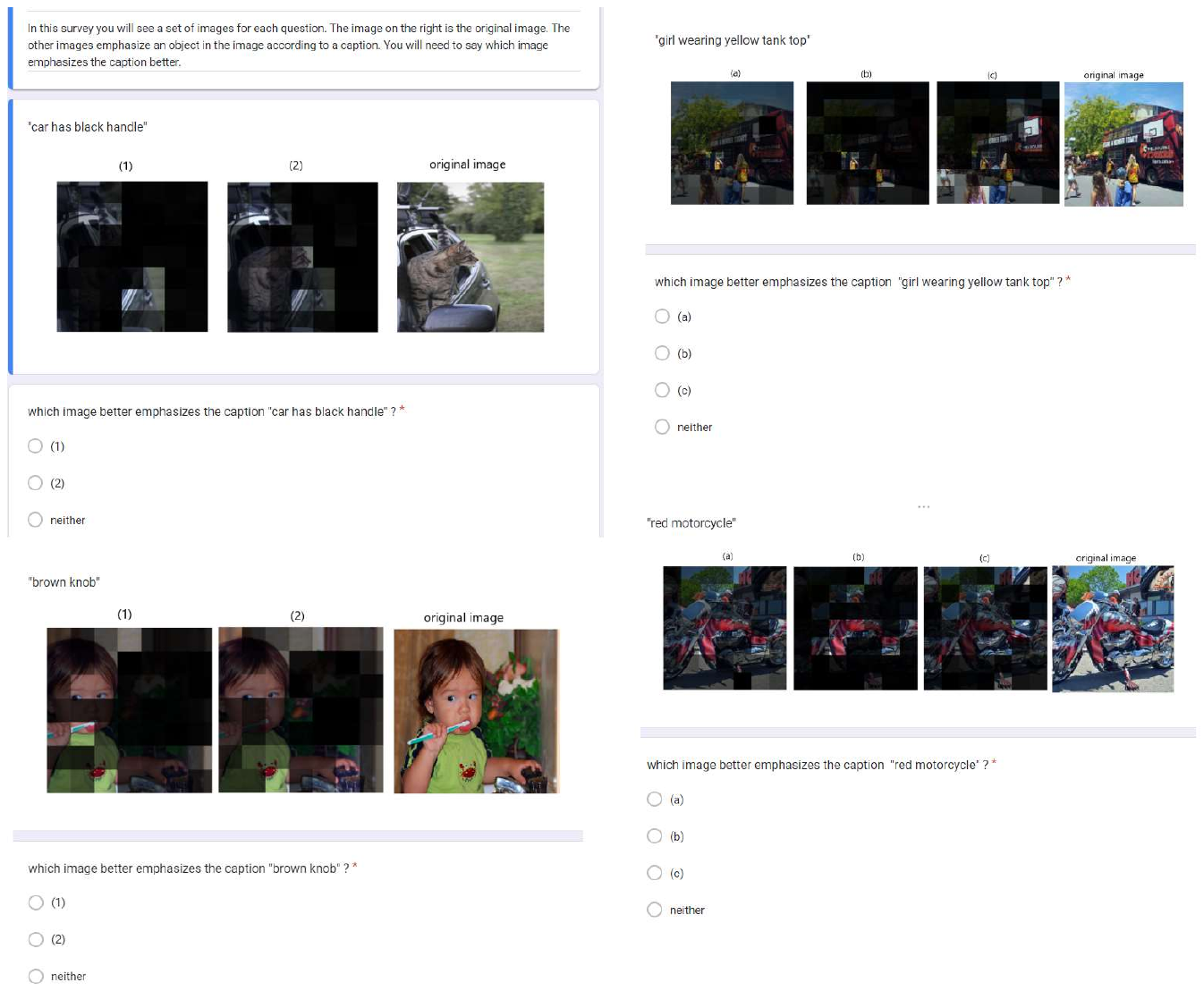}
\end{adjustbox}
\vspace{-1.7cm}
\caption{A screenshot of some questions from the user study. The user-study instructions appear at the top left.}
\label{fig:study}
\vspace{-0.5cm}
\end{figure*}

\appendices

\section{User Study Screenshots}

In Fig~\ref{fig:study} we provide several screenshots from our user study. They demonstrate the way the images with the questions were presented to the users in order to measure the interpretability of the different approaches.

\begin{IEEEbiography}  [{\includegraphics[width=1in,height=1.25in,clip,keepaspectratio]{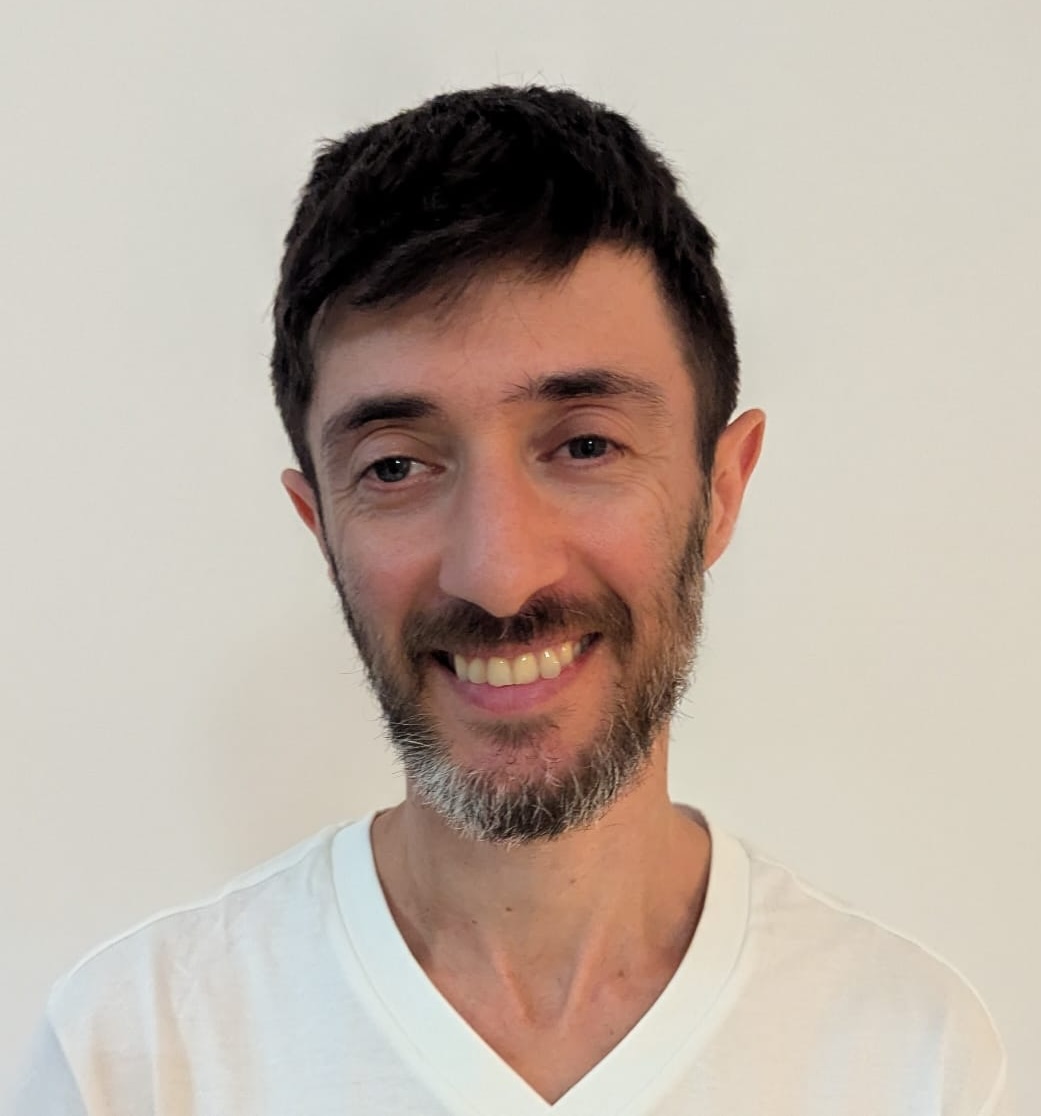}}]{Nir Yellinek} is a M.Sc graduate from Tel-Aviv University. He received his B.Sc (2011) and M.Sc. (supervision by Prof. Raja Giryes, 2023) degrees from the school of electrical engineering at Tel-Aviv University. His research interests include deep learning, self-supervised, multi-modal and AI alignment.
\end{IEEEbiography}

\begin{IEEEbiography}  [{\includegraphics[width=1in,height=1.25in,clip,keepaspectratio]{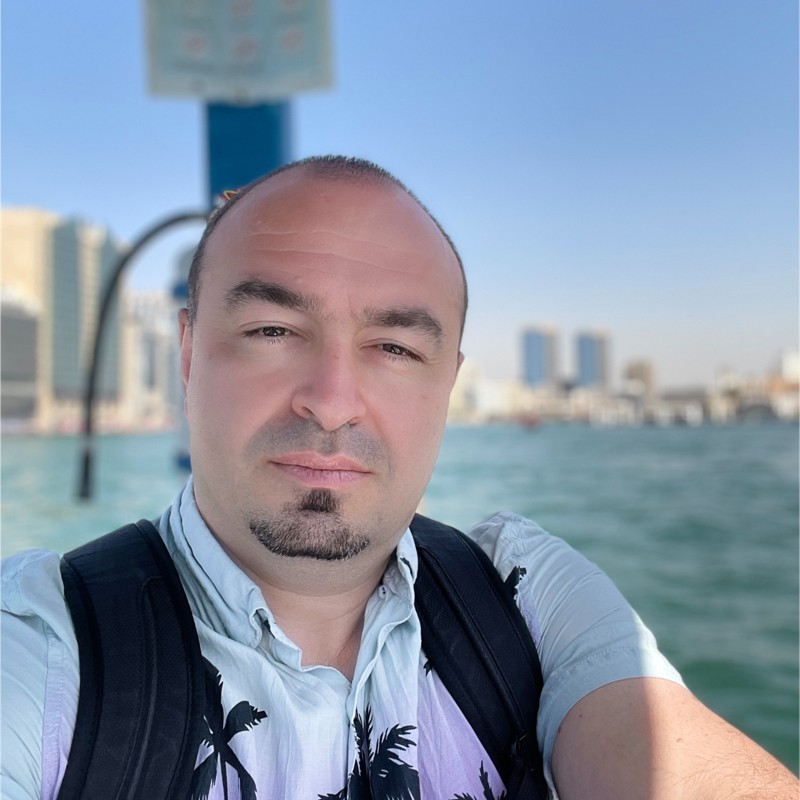}}]{Leonid Karlinsky}  is a Principal Research Scientist (STSM) in the MIT-IBM lab. Prior to that Leonid led the AI Vision research group in the Multimedia department @ IBM Research AI. Leonid joined IBM Research in July 2015. Before joining IBM, he served as a research scientist in Applied Materials, Elbit, and FDNA. He is actively publishing, reviewing, and performing occasional chair duties at ECCV, ICCV, CVPR, ICLR, AAAI, WACV, and NeurIPS, and is serving as an IMVC steering committee member for the past 6 years. During his time at IBM, Leonid has co-authored over 25 research papers on his research is in the areas of augmented reality, medical applications, self-supervised, cross-domain, multi-modal, and few-shot learning. He received his PhD degree at the Weizmann Institute of Science, supervised by Prof. Shimon Ullman.
\end{IEEEbiography}

\begin{IEEEbiography}  [{\includegraphics[width=1in,height=1.25in,clip,keepaspectratio]{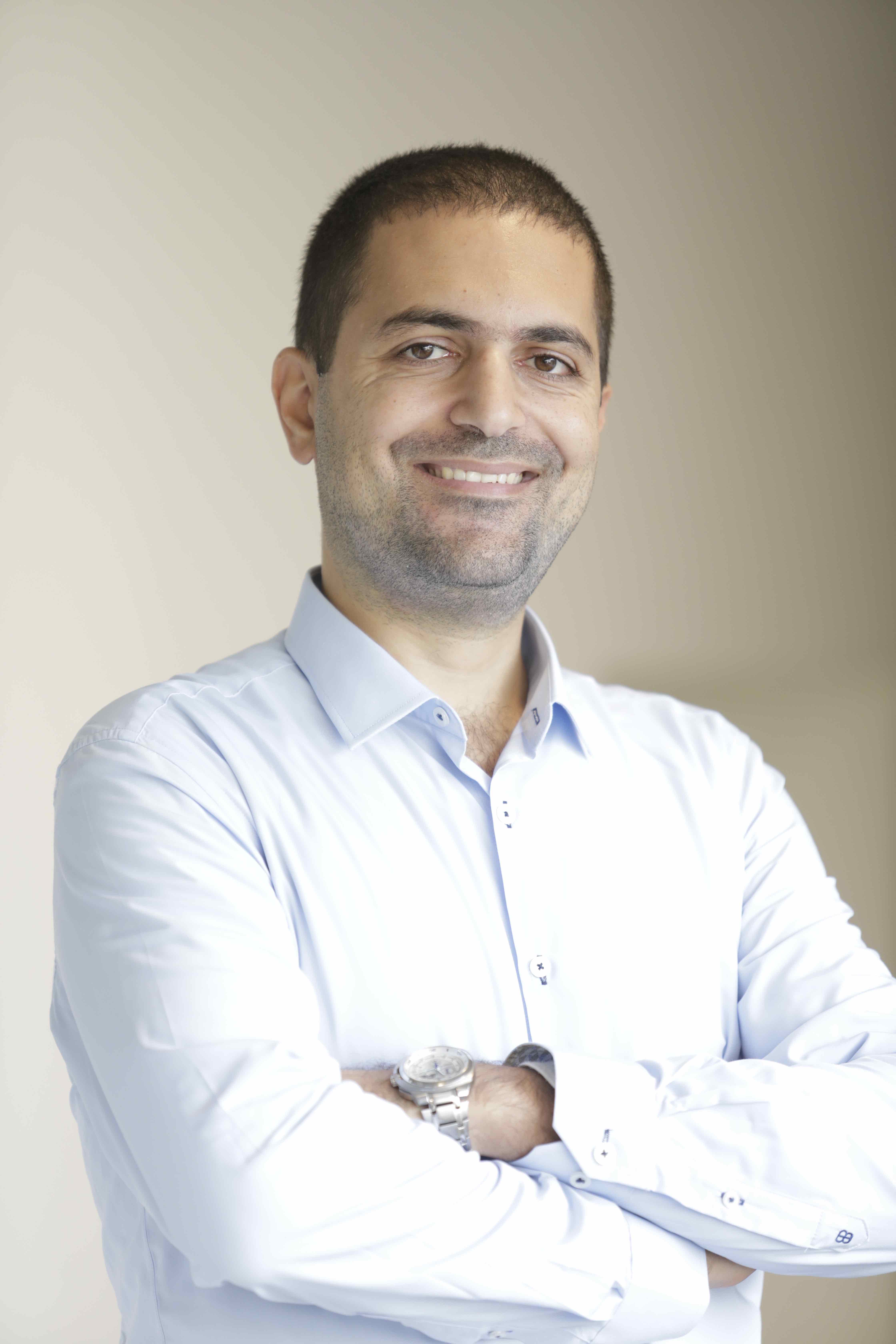}}]{Raja Giryes} is an associate professor in the school of electrical engineering at Tel Aviv University. His research interests lie at the intersection between deep learning and signal processing. Raja received the EURASIP best P.hD. award, the ERC-StG grant, Maof young faculty fellowship (2016-2019), VATAT postdoctoral scholarship (2014-2015), the TI Excellence in Signal Processing Award (ESPA)(2008), the Azrieli Fellowship (2010-2013). He is IEEE Senior Member and member of the Israeli young academy. He has 150+ publications in top venues with 9500+ citations. 30 MSc and 6 PhD students already graduated from his group. Two of them hold faculty positions at Bar Ilan University and University of Chicago. He is an associate editor in IEEE TPAMI, IEEE TIP and Elsevier Pattern Recognition and an area chair in NeurIPS. He organized workshops and tutorials on various aspects of deep learning both internationally and locally including in ICML, CVPR, ECCV and ICCV
\end{IEEEbiography}

\end{document}